# TurboAgent: An LLM-Driven Autonomous Multi-Agent Framework for Turbomachinery Aerodynamic Design

Juan Du(杜娟)[1,2,3], Yueteng Wu(吴跃腾)[1,2,3], Pan Zhao(赵攀)[1,2,3,*], Yuze Liu(刘裕泽)[4], Min Zhang(张敏)[1,2], Xiaobin Xu(徐晓斌)[1,2], Xinglong Zhang(张兴龙)[1,2,*]

[1]Digital Twin Research Center, Institute of Engineering Thermophysics, Chinese Academy of Sciences, Beijing 100190, China

[2]National Key Laboratory of Science and Technology on Advanced Light-duty Gas-Turbine, Chinese Academy of Sciences, Beijing 100190, China

[3]University of Chinese Academy of Sciences, Beijing 100190, China

[4]The University of Hong Kong, Hong Kong 999077, China

*Corresponding authors: zhaopan@iet.cn; zhangxinglong@iet.cn

## ABSTRACT

The aerodynamic design of turbomachinery is a complex and tightly coupled multi-stage process. It involves several key steps, including requirement analysis, geometry generation, performance prediction, optimization decision-making, and high-fidelity physical validation. Most existing studies on intelligent design focus on a single stage of the design process or connect disparate tools through customized scripted pipelines. Consequently, achieving a fully autonomous end-to-end design workflow remains challenging. To address this limitation, this study proposes TurboAgent, a large language model (LLM)-driven autonomous multi-agent framework for turbomachinery aerodynamic design and optimization. Within this framework, the LLM serves as the core for high-level planning and task coordination, while specialized agents are responsible for generative design, rapid performance prediction, multi-objective optimization, and high-fidelity physics validation. The proposed framework transforms the traditionally expert-dependent and trial-and-error-intensive iterative design process into a data-driven collaborative agent workflow, retaining high-fidelity numerical simulations as the definitive verification of physical consistency. A transonic single-rotor compressor is used as the validation case to systematically assess the proposed framework. The results demonstrate strong agreement among the design targets, generated configurations, and high-fidelity CFD simulations. The $R^2$ values for mass flow rate, total pressure ratio, and isentropic efficiency are all above 0.91, while the nRMSE values are below 8%. Furthermore, the meta-prompt-driven optimization agent improves the isentropic efficiency by 1.61% and the total pressure ratio by 3.02% relative to the initial design. These findings indicate that TurboAgent can autonomously execute the complete closed-loop design process, from natural language requirement inputs to final design generation. Under parallel computing, the entire workflow can be accomplished in approximately 30 minutes. TurboAgent provides an efficient and scalable new paradigm for turbomachinery aerodynamic design. It promotes a shift from conventional experience-based and simulation-driven iterative design toward an autonomous workflow with unified LLM planning and collaborative multi-agent execution.

# 1. INTRODUCTION

Turbomachinery is a core component in energy and power systems, such as aero-engines and gas turbines. Its aerodynamic performance and structural characteristics directly affect system efficiency, stability, and operational reliability. Driven by the growing demand for highly efficient energy conversion, reduced carbon emissions, high thrust-to-weight ratios, and compact configurations, the development of high-performance turbomachinery faces increasingly severe engineering challenges [1]. In particular, under high loading and high total pressure ratio, the three-dimensional (3D) aerodynamic design of turbomachinery blades has become exceedingly complex. The introduction of advanced design features, such as sweep and bend, further increases design freedom and significantly intensifies the nonlinear coupling among variables. As the dimensionality of the design space continues to grow, traditional design approaches that rely on expert experience and repeated trial and error are becoming less capable of meeting the engineering demands for high efficiency, superior quality, and rapid iteration.

Historically, turbomachinery design has followed a forward design paradigm dominated by engineering experience. Designers typically perform one-dimensional (1D) mean-line design, two-dimensional (2D) throughflow design, and 3D blade design sequentially to derive an initial geometry. This geometry is then refined through multiple rounds of numerical simulation and parameter optimization until the target performance is gradually approached [2]. In contrast, inverse design methods start from target performance metrics and directly derive the corresponding blade geometry, thereby reducing the reliance on flow simulations and partially shortening the design cycle. In this process, surrogate models coupled with high-dimensional optimization algorithms, such as genetic algorithms (GA) and particle swarm optimization (PSO), have further improved the search capability in complex design spaces [3-6]. However, these approaches still largely operate within an "expert-led, tool-assisted" paradigm. Firstly, they generally rely on manual definition of design objectives, constraints, and search strategies. Furthermore, under high-dimensional, multi-objective, and multidisciplinary coupling scenarios, they still require extensive simulation, expert judgment, and repeated iteration. As a result, long design cycles, high computational costs, and the limited explicit reuse of knowledge remain significant challenges.

In recent years, data-driven methods have opened a new pathway for turbomachinery design. The rapid development of deep learning (DL) has created new opportunities for engineering design [7]. Generative models, such as generative adversarial networks (GANs) [8], variational autoencoders (VAEs) [9], and diffusion probabilistic models (DPMs) [10], can learn high-dimensional nonlinear mappings between geometries and performance metrics. Consequently, they can rapidly generate candidate designs under given performance requirements and design constraints. Integrated with other data-driven techniques, these methods also alleviate the reliance on high-fidelity simulations and expert experience [11]. This synergy has significantly enhanced the efficiency of generating and screening candidate designs, promoting a gradual transition in turbomachinery design from conventional experience-driven practices to a data-driven paradigm. Related studies have achieved encouraging results in fields such as aerospace engineering and structural design [12-13], and have also created new opportunities for aerodynamic inverse design of turbomachinery. More recently, LLMs, as an important form of general artificial intelligence, have demonstrated strong capabilities in reasoning, planning, natural language understanding, and cross-tool coordination [14]. Existing studies have successfully applied them to tasks including code generation [15], CAD modeling [16-17], and simulation analysis [18]. This progress suggests that LLMs may overcome the limitations of using a single model for a single task, and instead organize multiple design stages into a higher-level intelligent workflow.

Despite these advances, the majority of existing intelligent design research remains focused on isolated stages, such as geometry generation, performance prediction, or optimization. Even when efforts attempt to bridge disparate tools via scripts or predefined pipelines, they essentially manifest as rigid, function-level integrations rather than autonomous systems capable of dynamic adaptation based on task status and execution feedback. For turbomachinery aerodynamic design, localized intelligence or static coupling is fundamentally insufficient to meet practical engineering demands. The fundamental reason is that turbomachinery aerodynamic design is inherently a complex process with multiple stages, strong coupling, and heavy dependence on feedback. It involves several key steps, including requirement analysis, geometry generation, performance prediction, optimization decision-making, and high-fidelity physical validation. These stages exhibit clear hierarchical relationships and iterative dependencies. Therefore, the critical unmet need is not merely a more robust model for an individual task, but an autonomous framework capable of unified cross-stage planning, dynamic collaborative execution, and closed-loop validation via high-fidelity numerical simulations. Concurrently, LLM-driven multi-agent systems (MAS) have shown promising results in materials design [19], autonomous experimental planning [20], software debugging [21], and scientific computation [22]. These breakthroughs indicate the potential of this paradigm for automated coordination and closed-loop optimization in complex engineering problems. However, its application to turbomachinery aerodynamic design remains largely unexplored.

To address the above challenges, this study proposes TurboAgent, an LLM-driven autonomous multi-agent framework for turbomachinery aerodynamic design and optimization, as illustrated in **Fig. 1**. Unlike existing studies that mainly focus on single-condition pointwise generation, localized prediction, or scripted simulation calls, TurboAgent transcends the mere embedding of AI models into a conventional design process. Instead, it reformulates turbomachinery aerodynamic design into an end-to-end autonomous workflow. Within this framework, the LLM performs unified planning and coordination, multiple functional agents execute specialized tasks collaboratively, and high-fidelity physical validations serve as the ultimate constraint for consistency. Acting as the central orchestrator, the LLM translates natural-language design requirements into structured objectives and executable procedures, dynamically updating the task plan based on execution feedback. Meanwhile, the functional agents are responsible for generative geometry design, rapid performance prediction, multi-objective optimization, and physics-based validation using computational fluid dynamics (CFD) and finite element analysis (FEA). This collaborative mechanism seamlessly links requirement interpretation, candidate generation, performance prediction, optimization search, and physical verification. Consequently, TurboAgent transforms the traditional expert-dependent, trial-and-error iterative design process into a data-driven, agent-based workflow, while retaining high-fidelity numerical simulation as the final means of physical consistency verification.

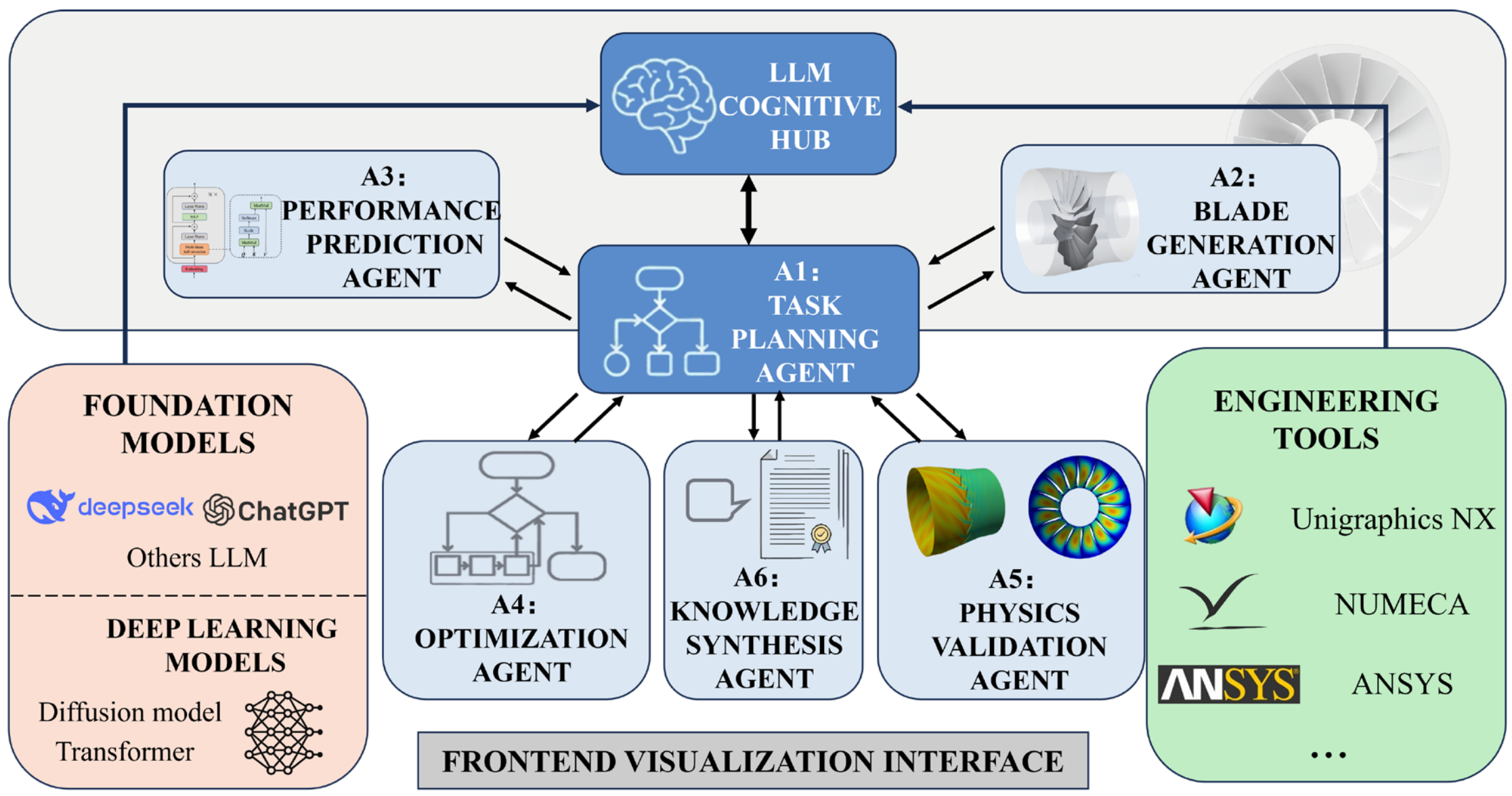

Figure 1. Multi-agent AI framework for turbomachinery design

The main contributions of this study are summarized as follows:

(1) An LLM-driven autonomous multi-agent framework is proposed for turbomachinery aerodynamic design. This framework enables the unified orchestration of requirement interpretation, task planning, geometry generation, performance prediction, optimization decision-making, and high-fidelity physical validation.

(2) A synergistic design mechanism is established, integrating data-driven rapid generation with high-fidelity physical validation. Specifically, a conditional denoising diffusion probabilistic model (cDDPM) is used for high-dimensional inverse geometry generation, a Transformer-based surrogate model is adopted for rapid performance prediction, and LLM-driven multi-objective optimization is integrated with automated CFD/FEA validation. Together, these components improve both design iteration efficiency and engineering applicability.

(3) The effectiveness and engineering feasibility of the proposed framework are systematically validated through a transonic single-rotor compressor case. The evaluation covers multiple aspects, including task planning, single-agent capability, end-to-end closed-loop workflow, and computational cost. The results demonstrate that the framework can support autonomous turbomachinery design driven by natural-language requirements.

The remainder of this paper is organized as follows. Section 2 reviews recent advances in generative AI and LLM-based agents for engineering design. Section 3 presents the architecture and functional roles of the proposed multi-agent framework. Section 4 demonstrates the effectiveness of the proposed framework through a transonic single-rotor compressor case. Finally, Section 5 concludes the paper and outlines directions for future work.

## 2. Related works

This section reviews recent advances in generative models for aerodynamic design and summarizes the development of LLM-based agent systems in engineering design. The review establishes the foundation for understanding how the proposed autonomous multi-agent generative design framework addresses existing gaps in turbomachinery aerodynamic design.

### 2.1 Generative models for aerodynamic shape design

Generative artificial intelligence (GAI) enables the direct mapping between target performance and geometric parameters, making it a powerful tool for aerodynamic design. In 2D airfoil design, Wang et al. proposed Airfoil-GAN, achieving airfoil reconstruction and generation in the latent space [23]. Achour et al. employed a cGAN to construct models based on angle-of-attack or lift-to-drag parameters, enabling efficient optimization under flow constraints [24]. More recently, DPMs have demonstrated stronger generation capability and improved constraint handling. Wei et al. proposed DiffAirfoil, which enables high-quality airfoil sampling using limited training data [25]. Wen et al. introduced Airfoil-DDPM, which supports controllable airfoil generation under multiple performance objectives through conditional encoding [26].

For turbomachinery design, applications of generative models can be broadly categorized into two directions. First, generative models have been employed as efficient surrogate models to accelerate high-fidelity numerical simulations [27-28]. Representative examples include cGAN-based pressure field prediction [27] and temperature field prediction for turbine film-cooling hole design [29]. Joachim et al. combined a VAE with gated recurrent units (GRUs) to capture unsteady flow features in high-pressure turbines, including shock interactions and wake evolution [30]. Second, generative models have been increasingly applied to the direct inverse design of turbomachinery [31-34]. Ghosh et al. combined Gaussian processes with conditional invertible neural network (CINN) to develop a probabilistic inverse design framework for gas turbine blades [32]. Wu et al. introduced a framework combining DPMs with gradient-based optimization to generate 3D compressor blades from performance targets [33], and further introduced a prior-guided diffusion model for single-stage turbine generation [34]. Guo et al. developed a sample-weighted VAE to enable knowledge transfer across operating conditions in the latent space, thereby improving optimization efficiency in new scenarios [35].

Despite these advances, most existing methods remain model-driven generation approaches and lack the capability to dynamically interpret design intent. In addition, geometry generation and high-fidelity simulation are usually connected through scripts. This limits the integration of physical knowledge and makes autonomous closed-loop design difficult.

### 2.2 LLMs and agentic AI in engineering design

In recent years, LLMs have experienced rapid adoption in industrial and mechanical engineering design, particularly in multimodal knowledge acquisition, text-driven geometric modeling, and multi-agent collaborative workflows.

In the area of knowledge acquisition and geometric modeling, Li et al. proposed MechRAG, which integrates CAD, CAE, and technical documentation to improve the accuracy of complex design question-answering tasks [36]. Hu et al. developed a multimodal knowledge graph to support intelligent question answering for wind turbine assembly processes [37]. For text-to-CAD tasks, Sun et al. introduced LLM4CAD, which translates natural language descriptions into executable CAD command sequences [16]. Subsequent studies further incorporated sketches and view information as additional conditions to improve generation accuracy [17].

In conceptual design and optimization decision-making, LLMs demonstrate expert-level reasoning capabilities. Jiang et al. proposed AutoTRIZ, integrating TRIZ theory with LLMs to facilitate contradiction identification and solution generation [38]. Yao et al. demonstrated that LLM-generated aircraft conceptual designs can achieve quality comparable to those of human engineers [39]. In optimization tasks, LLMs can understand design semantics and autonomously adjust search strategies, with successful applications reported in robotic and structural design domains [40-42]. Jadhav et al. [43] and Zhang et al. [44] further demonstrated the feasibility of LLMs in 2D structural generation and parametric shape optimization. Using the RocketBench platform, Simonds et al. showed that reinforcement learning-enhanced LLMs can outperform human experts in rocket design tasks [45].

For complex engineering tasks such as materials design, drug discovery, and autonomous experiment planning, agentic AI frameworks provide efficient solutions. Compared with standalone LLMs, these frameworks decompose design processes into subtasks using role allocation, state management, and tool invocation, enabling multi-round collaboration, knowledge sharing, and decision optimization [46-47]. For example, Kang et al. developed a natural language-driven platform integrating LLMs, domain-specific databases, and machine learning models for the design of metal-organic frameworks [19]. Alireza et al. proposed ProtAgents, an LLM-based protein design system in which multiple agents collaboratively complete design tasks while integrating knowledge retrieval and simulation [48]. In the CFD domain, researchers have developed a CFD Agent capable of mesh generation, solver configuration, and post-processing through natural language interaction [49]. Elrefaie et al. proposed a framework combining LLMs and deep learning to generate 3D vehicle models from 2D sketches and conduct aerodynamic analysis, significantly accelerating vehicle design [50]. These studies demonstrate that integrating LLMs with agentic AI enhances cross-domain task planning, knowledge reasoning, and process transparency. This paradigm provides a promising pathway for turbomachinery aerodynamic design.

Overall, existing research on LLMs and MAS shows potential in knowledge acquisition, geometric generation, and design reasoning. However, a systematic framework specifically tailored to turbomachinery aerodynamic design is still lacking. Therefore, developing an autonomous design framework centered on LLMs and tightly coupled with generative models and simulations is crucial for advancing aerodynamic design from "model-assisted" to "autonomous intelligent" paradigms.

# 3. Methods

## 3.1 System framework design

### 3.1.1 Overall architecture overview

TurboAgent is built on the LangGraph framework (LangChain Team, 2024), a graph-structured agent framework that enables efficient integration of LLMs with external tools, APIs, and databases. This architecture supports the development of complex intelligent systems for engineering design. The core of LangGraph lies in its graph-based representation, where nodes correspond to agents or tasks, and edges represent dependencies or information flow between them. Such a structure enables flexible management of complex workflows and supports dynamic task scheduling in response to evolving design requirements.

The proposed framework adopts a supervisor-based architecture. A high-level planning agent is responsible for global task scheduling and decomposes complex turbomachinery aerodynamic design problems into multiple subtasks. Each agent is designed to handle a specific stage of the design process, including task interpretation, blade generation, performance prediction, optimization decision-making, and physical validation. Through the collaboration of multiple agents, the overall engineering design workflow is significantly simplified and the design efficiency is improved.

The workflow of the proposed framework is illustrated in **Fig. 2**. First, the designer specifies design requirements and constraints through natural language input. The task planning agent interprets user intent and decomposes it into a dynamically executable workflow. Subsequently, the blade generation agent produces candidate geometries and performs visualization, while the performance prediction agent rapidly evaluates multiple design schemes. Based on the prediction results and expert feedback, the task planning agent determines whether the optimization agent should be activated. During the optimization stage, the optimization agent dynamically adjusts the search strategy according to design objectives. After optimization, candidate designs are further validated through high-fidelity physical simulations, including CFD simulations and structural strength analysis. The physics validation agent automatically post-processes the simulation results, extracts useful design information, and identifies the optimal design scheme. Once the design task is completed, the system generates a comprehensive design report.

The entire design process is dynamically scheduled by the high-level task planning agent. Depending on the design requirements, different agents can operate independently or collaboratively. This significantly improves design efficiency, reduces manual intervention, and enhances both the diversity and accuracy of the generated design schemes. In addition, human feedback can be introduced as external input to guide the agents in handling ambiguous design decisions. This reduces the number of design iterations and the dependence on domain experts, providing a scalable approach for autonomous turbomachinery design.

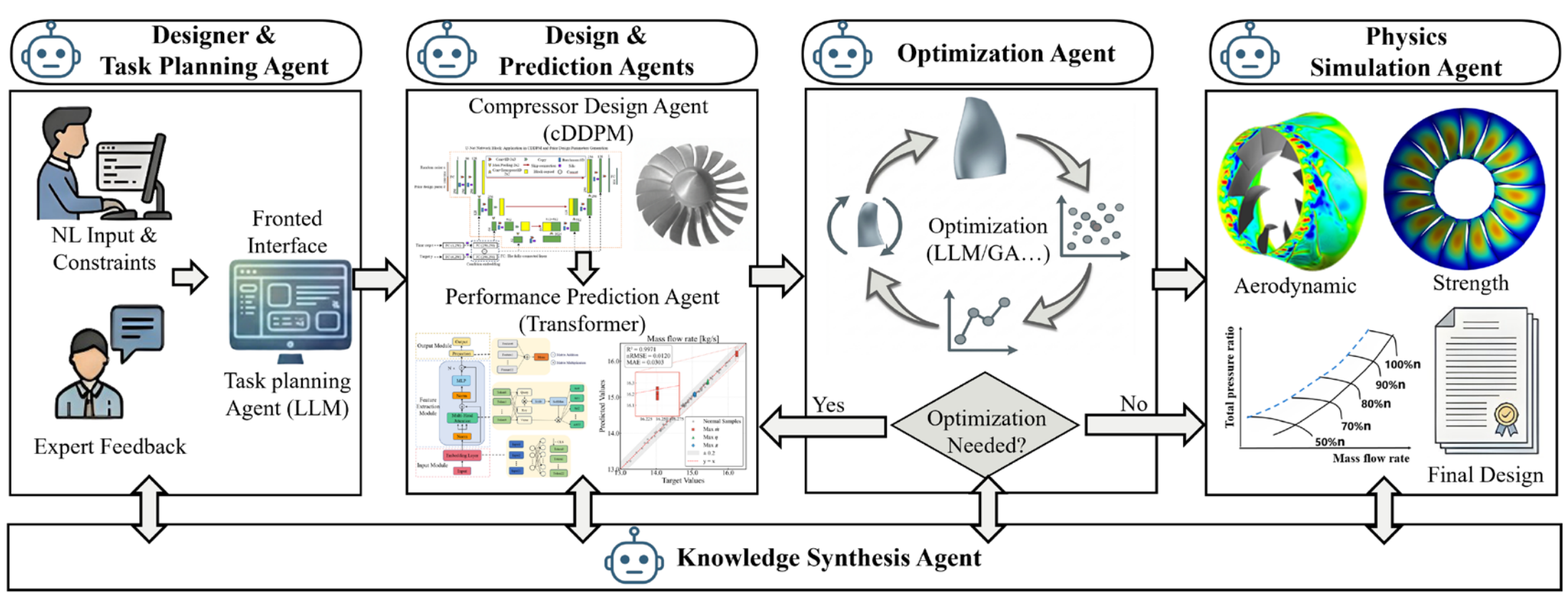

Figure 2. The workflow of TurboAgent

### 3.1.2 Database construction and parametric design

Within a multi-agent collaborative design framework centered on generative models, the construction and parameterization of the compressor database play a critical role in the overall system. The database used in this study, derived from [33], is based on a prototype of a transonic 1.5-stage compressor rotor blade. In turbomachinery design and optimization, parameterization is a fundamental step, as it enables the representation of complex geometries using a finite set of predefined design variables.

As illustrated in Fig. 3, the 3D blade geometry is constructed by stacking multiple 2D airfoil sections along the spanwise direction, from hub to tip, based on their centroids. Each section profile is generated by superimposing a thickness distribution onto a camber line. Both the camber line and the thickness distribution are described using Non-Uniform Rational B-splines (NURBS).

In this study, parameterization is applied to three representative sections: the hub (*span* = 0.0), mid-span (*span* = 0.5), and tip (*span* = 1.0). For each section, seven independent design variables are defined while keeping other parameters fixed, resulting in a total of 21 design variables. These variables include: leading-edge metal angle ($\beta_{1k}$), trailing-edge metal angle ($\beta_{2k}$), chord length (*C*), maximum relative thickness (*T*), position of maximum relative thickness (*TP*), rotor bend (*RB*), and rotor sweep (*RS*). The bend and sweep are controlled by translating each section along the directions perpendicular and parallel to the chord, respectively. Detailed descriptions of the parameterization method can be found in our previous work [33].

This parameterization scheme enables the efficient representation of the full 3D blade geometry using a limited number of physically meaningful variables. Such representation facilitates the incorporation of LLM-based knowledge into the design process. Furthermore, the constructed database provides essential support for training both the generative design agent and the performance prediction agent within the proposed framework.

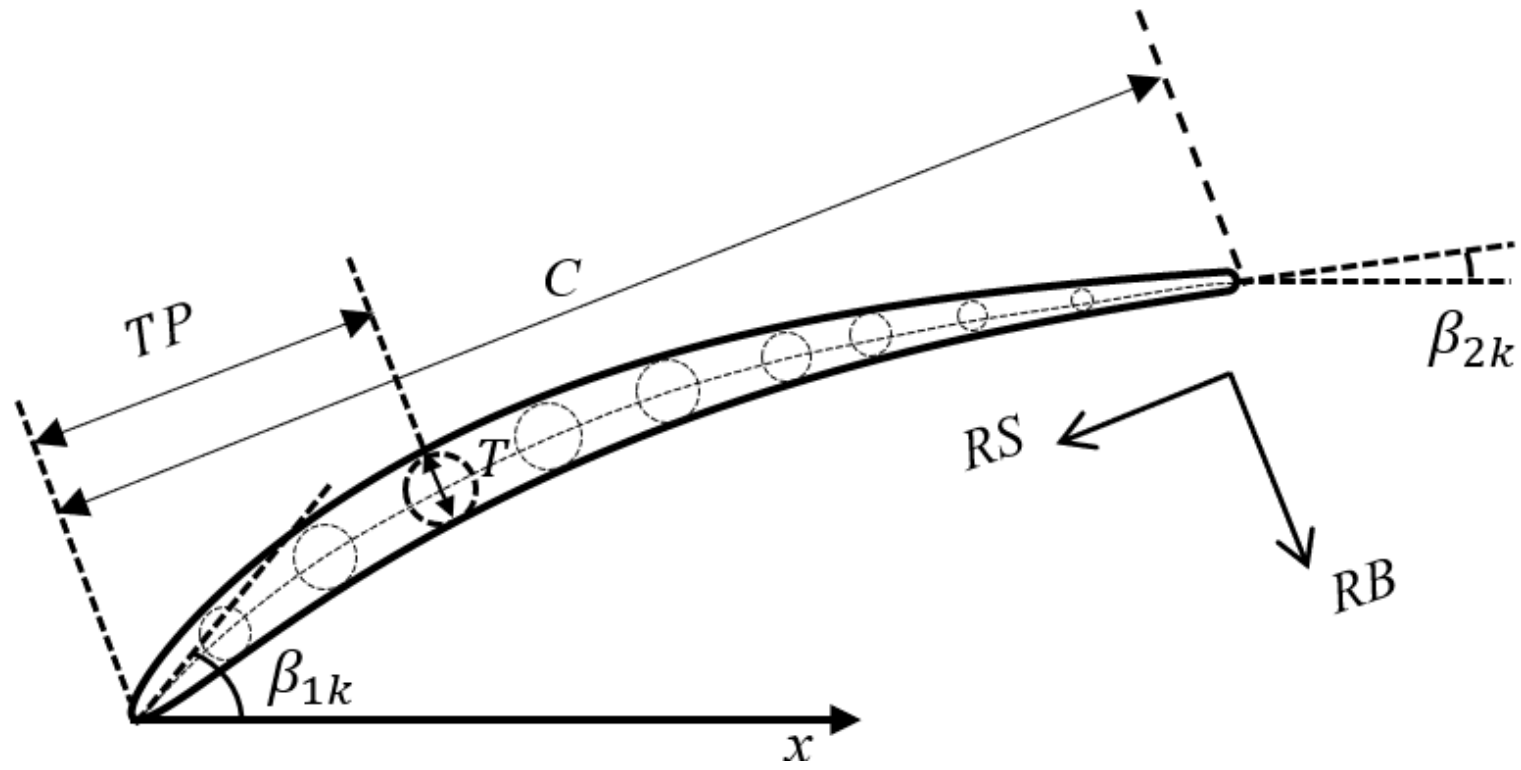

Figure 3. Two-dimensional blade profile parameterization.

### 3.1.3 Front-end interface design

As shown in Fig. 4, a front-end interface is developed to improve user experience and enhance the transparency of the design workflow. The interface is built using Flask, a lightweight web framework that supports rapid application development and deployment. The front-end integrates the MAS into a unified platform.

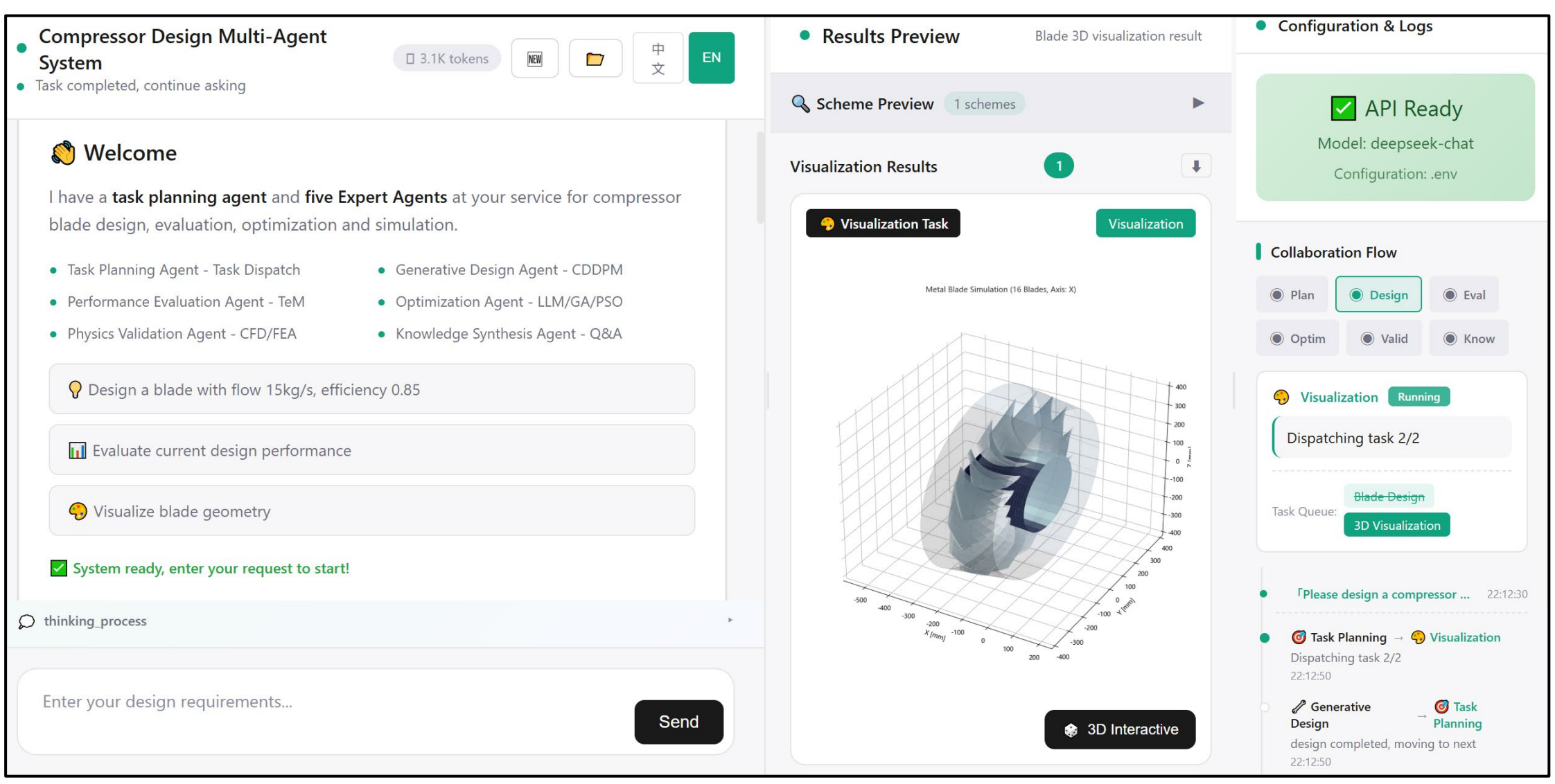

Figure 4. Front-end interface of the TurboAgent system.

Communication between the front-end and the back-end agent system is implemented through RESTful APIs, enabling efficient interaction and providing an intuitive graphical user interface (GUI). Designers can specify design requirements using natural language, visualize generated design schemes, adjust key parameters, and receive real-time feedback from the agents, including optimization results.

The interface further supports the visualization of geometric configurations, performance predictions, and simulation outcomes, allowing users to comprehensively evaluate the feasibility and effectiveness of the generated designs. In addition, a post-processing module is incorporated into the system, which automatically extracts key information from simulation data and presents the results graphically, thereby facilitating informed design decision-making.

The adoption of the Flask framework ensures efficient coordination between the front-end and back-end components, while also providing flexibility and scalability for future system extensions.

### 3.2 Task planning agent

The high-level task planning agent serves as the core component of the proposed framework. It is responsible for global task orchestration, including task parsing, intent recognition, task scheduling, and decision routing. Based on design requirements and intermediate results, the agent dynamically updates the task plan to ensure both flexibility and robustness throughout the design process.

By leveraging different LLM APIs, the agent interprets natural language inputs provided by the user. It combines semantic understanding with rule-based matching to convert user requests into structured task sequences. During execution, the agent continuously refines design decisions based on real-time feedback.

By dynamically adjusting task order and priority, the high-level planning agent improves overall design efficiency and adaptability. It can also selectively invoke the compressor design agent, performance prediction agent, and optimization agent at appropriate stages. This enables iterative refinement of design solutions and ensures convergence toward optimal results across multiple design cycles. The detailed prompt design and instruction templates of the task planning agent are provided in Appendix A.1.

### 3.3 Generative design agent

As a core component of the proposed multi-agent system, the generative design agent enables efficient geometry generation conditioned on target performance ($\dot{m}$, $\pi$, and $\eta$) by leveraging a conditional denoising diffusion probabilistic model. Compared with GANs and VAEs, DPMs progressively corrupt data with noise and learn the reverse denoising process. This mechanism produces higher-quality samples with finer details and avoids training instability, making it particularly suitable for complex engineering design tasks.

DPMs are probabilistic generative models consisting of two stages: a forward diffusion process and a reverse denoising process. In the forward process, noise is gradually added to a data sample. In the reverse process, a neural network is trained to remove noise and recover the original data. The training and sampling procedures are summarized in Table 1.

In the forward diffusion process, given a real data sample $x_0$, noise is progressively added to generate a sequence $x_1, x_2, \ldots, x_T$, where $T$ is the total number of diffusion steps. Each step is defined as:

$$q(x_t \mid x_{t-1}) = \mathcal{N}(x_t; \sqrt{1-\beta_t} x_{t-1}, \beta_t I) \tag{1}$$

where $\beta_t$ is the noise schedule controlling the noise magnitude at each step, and $\mathcal{N}$ denotes a Gaussian distribution. As $t$ increases, the data gradually becomes corrupted and eventually approaches pure noise $x_T$.

The reverse process learns a denoising model $\epsilon_\theta$ to recover the original data from noise. The model is trained by minimizing the difference between the predicted noise and the true noise:

$$L = \mathbb{E}_{\boldsymbol{x}_0,\epsilon,t,\boldsymbol{c}}[\| \epsilon - \epsilon_\theta(\sqrt{\bar{\alpha}_t}\boldsymbol{x}_0 + \sqrt{1-\bar{\alpha}_t}\epsilon, t, \boldsymbol{c})\|^2] \tag{2}$$

where $\epsilon \sim \mathcal{N}(0, \mathbf{I})$, and $c$ represents the conditional input. The model parameters $\theta$ are optimized using mean squared error (MSE) during training.

In terms of model architecture, a 1D U-Net is adopted as the backbone of the denoising network to capture the sequential characteristics of parameterized compressor blade data [34]. The architecture consists of an encoder, a decoder, and skip connections, which enable the effective fusion of multi-scale geometric features.

To further improve generation accuracy and convergence, a deep neural network (DNN)-based guidance mechanism is introduced, following our previous work [34]. This mechanism constructs a hybrid conditional vector by jointly encoding physical design targets and geometric prior information. It provides directional guidance for the diffusion process, thereby enhancing both generation efficiency and accuracy.

In addition, to facilitate practical engineering applications, the generated geometric parameters are further converted into 3D compressor blade models using the Siemens NX (UG) platform. This process enables automated parametric modeling and provides standardized geometry inputs for subsequent high-fidelity structural strength analyses. As a result, an effective link is established between data-driven design and physics-based engineering validation.

**Table 1. Training and sampling algorithms of the cDDPM-based generative model.**

| **Algorithm 1:** Conditional diffusion model: training and sampling |
|---|
| 1: **Phase 1: Training** |
| 2: **repeat** |
| 3: sample data $x_0 \sim q(x_0)$, condition $c \sim q(c)$ |
| 4: sample time step $t \sim \text{uniform}(\{1,...,T\})$ |
| 5: sample noise $\epsilon \sim \mathcal{N}(0, I)$ |
| 6: perform gradient descent step on: $\nabla_\theta \left\| \epsilon - \epsilon_\theta\left(\sqrt{\bar{\alpha}_t}x_0 + \sqrt{1-\bar{\alpha}_t}\epsilon, t, c\right)\right\|^2$ |
| 7: **until** converged |
| 8: **Phase 2: Sampling** |
| 9: Given condition $c$, sample $x_T \sim \mathcal{N}(0, I)$ |
| 10: **for** $t = T,...,1$ **do** |
| 11: sample $z \sim \mathcal{N}(0, I)$ if $t > 1$, else $z = 0$ |
| 12: update: $x_{t-1} = \frac{1}{\sqrt{\alpha_t}}\left(x_t - \frac{1-\alpha_t}{\sqrt{1-\bar{\alpha}_t}}\epsilon_\theta(x_t, t, c)\right) + \sigma_t z$ |
| 13: **end for** |
| 14: **reture** $x_0$ |

### 3.4 Performance prediction agent

The performance prediction agent is designed to provide rapid evaluation of generated design candidates. It can also collaborate with the optimization agent to support iterative design improvement. In this study, the agent adopts the surrogate model developed in [33], which is based on a Transformer-encoder architecture. This model enables real-time prediction of key aerodynamic performance metrics, including $\dot{m}$, $\pi$, and $\eta$. The input-output relationship is defined in Eq. (3).

$$\dot{m}, \pi, \eta = f(x_1, x_2, \ldots, x_{21}) \tag{3}$$

Since its introduction in 2017, the Transformer architecture has achieved remarkable success across multiple domains, including natural language processing, computer vision, speech recognition, and flow field prediction. Unlike conventional convolutional neural networks (CNNs) and recurrent neural networks (RNNs), the Transformer employs a self-attention mechanism to capture global dependencies among sequence elements, as shown in Eq. (4). This capability overcomes the limitations of CNNs and RNNs in modeling long-range dependencies.

$$\text{Attention}(\boldsymbol{Q}, \boldsymbol{K}, \boldsymbol{V}) = \text{softmax}\left(\frac{\boldsymbol{Q}\boldsymbol{K}^T}{\sqrt{d_k}}\right)\boldsymbol{V} \tag{4}$$

In turbomachinery design, aerodynamic performance is determined by complex and highly coupled relationships among multiple design parameters. The adopted Transformer encoder architecture is particularly well-suited to address this characteristic. This architecture enables the accurate and efficient prediction of compressor performance under diverse design conditions, thereby supporting rapid design evaluation and optimization.

### 3.5 Optimization agent

The optimization agent is a key module in the proposed system. It refines the initial designs generated by the generative agent to further improve specific performance objectives. This module integrates multiple optimization strategies, including an LLM-driven optimization approach based on meta-prompting, as well as conventional algorithms such as GA and PSO. GA performs global search by simulating natural selection and iteratively improving solutions through crossover and mutation. PSO achieves global optimization by mimicking collective search behavior, where particles update their positions based on shared information.

Motivated by the strong performance of LLMs in evolutionary search tasks, this study proposes an LLM-driven aerodynamic optimization method for turbomachinery. In this approach, the LLM is treated as an intelligent optimization agent with semantic understanding and adaptive exploration capabilities. Through prompt engineering, the model can perform multi-objective optimization without requiring gradient information or explicitly defined evolutionary operators.

The workflow of the proposed LLM-driven algorithm is summarized in **Table 2**. During the optimization stage, design variables, performance metrics, and constraint conditions are embedded into prompts in natural language form, explicitly defining the optimization objectives and evaluation criteria. Based on the prompt, the LLM interprets the task semantics and generates a new mean vector $\bar{x}$ for the next generation. It is assumed that the design vectors in each generation follow an $N$-dimensional Gaussian distribution: $x_i \sim \mathcal{N}(\bar{x}, \sigma^2 \mathrm{I})$. After obtaining the mean vector, $N$ candidate design samples are generated through Gaussian sampling. The variance is dynamically determined by the LLM during the optimization process, ensuring adaptive exploration of the design space.

**Table 2. Workflow of the LLM-driven optimization algorithm.**

| **Algorithm 2:** LLM-driven compressor aerodynamic performance optimization. |
|---|
| 1: **Input:** Design space $x_i \in \mathbb{R}^n$; objective functions $S_i$; constraints $C$; LLM hyperparameters (temperature, top p, etc.). |
| 2: **Output:** Global optimal design $x^*$ and corresponding performance $S_i$ |
| 3: Initialize the design database $D_0$ using Latin Hypercube Sampling (LHS) and evaluate aerodynamic performance. Construct the weighted objective function: $R = \frac{\sum_{i=1}^{K} w_i S_i}{\sum_{i=1}^{K} w_i} - \sum_{j=1}^{M} w_j P_j$ |
| 4: **for** generation $g \leftarrow 1$ to $G_{max}$ **do** |
| 5: Encode design variables and constraints into a natural language prompt for the LLM. |
| 6: LLM infers a mean vector $\bar{x}$. |
| 7: Generate candidate designs via Gaussian sampling: $x \sim N(\bar{x}, \sigma^2 I), i = 1, \ldots N$ . |
| 8: Evaluate aerodynamic performance $f_k(x_i)$ for all candidates. |
| 9: Update the LLM prompt and adapt $\sigma$ based on feedback. |
| 10: Update the historical best design if improvement is found. |
| 11: **if** convergence criteria are met (e.g., improvement smaller than $\varepsilon$ for $J$ consecutive generations) **then** |
| 12: **break;** |

The performance prediction agent then evaluates the aerodynamic performance of candidate designs. The results are ranked according to predefined criteria and fed back into the prompt. Through iterative updates and feedback, the optimization process progressively converges toward high-quality solutions. Convergence is assumed when the improvement in the best performance remains below a predefined threshold for $J$ consecutive generations. The detailed prompt template is provided in Appendix A.1.

Additionally, to handle multi-objective optimization, a unified reward function is defined as:

$$R = \frac{\sum_{i=1}^{K} w_i S_i}{\sum_{i=1}^{K} w_i} - \sum_{j=1}^{M} w_j P_j \qquad (5)$$

where $R$ is the overall reward, $S_i$ represents the score of the $i$-th objective, and $w_i$ is its corresponding weight. $P_j$ denotes the penalty associated with the $j$-th constraint, with weight $w_j$. $K$ and $M$ are the numbers of objectives and constraints, respectively.

For performance metrics such as $\dot{m}$, $\pi$, and $\eta$, values are first normalized based on the statistical range of the training data and mapped to a score range of [0,100]. For maximization objectives, the score is defined as $S = \hat{v} \times 100$, while for minimization objectives, it is defined as $S = (1 - \hat{v}) \times 100$, where $\hat{v}$ is the normalized value. When constraints are violated, penalties are computed as linear functions of the normalized deviation. If the design evaluation fails, a fixed penalty of $R = -1000$ is assigned. This reward formulation converts the multi-objective design problem into a scalar optimization problem. It should be noted that the weights of objectives and constraints are dynamically determined by the LLM based on design goals, evaluation results, and user-provided requirements.

### 3.6 Physics validation agent

The physics validation agent is responsible for the final verification of generated design candidates. It focuses on both aerodynamic performance and structural strength, which are critical for turbomachinery design. Aerodynamic validation ensures stable operation under different working conditions, while strength analysis guarantees mechanical reliability under loading. In conventional design workflows, physical validation requires extensive manual effort. This includes setting boundary conditions, generating computational meshes, configuring solvers, and performing post-processing through graphical interfaces. Such procedures are time-consuming and create a high barrier for practical use.

To improve simulation efficiency, the proposed framework integrates the LLM into a simulation workflow. Through natural language interaction, the LLM automatically configures solver settings, executes simulation tasks, and retrieves results, enabling a high level of automation.

For aerodynamic validation, the agent performs CFD simulations using domain-specific tools. Structured hexahedral meshes are automatically generated using NUMECA AutoGrid5 with an H-O-H topology. The flow field is solved using ANSYS CFX, employing the shear stress transport (SST) turbulence model. At the inlet, the total temperature and total pressure are specified under standard atmospheric conditions. Different operating points are obtained by adjusting rotational speed and outlet static pressure. This integration of LLM with domain tools significantly reduces manual intervention and improves simulation efficiency.

For structural validation, the agent employs a custom-developed FEA solver to perform strength analysis. Designers can modify material properties, such as Young's modulus, Poisson's ratio, and density, as well as geometric parameters through natural language input. This interaction improves both flexibility and usability in the design process.

The post-processing of simulation results is also automated. The agent extracts key physical quantities from the flow field and identifies important features such as shock structures and leakage flows. The results are visualized in the front-end interface, enabling rapid comparison and evaluation of different design schemes. This capability supports more efficient optimization decisions.

Overall, the physics validation agent tightly integrates LLM capabilities with simulation workflows. It automates solver configuration, reduces manual intervention, and accelerates the evaluation and optimization of turbomachinery designs.

### 3.7 Knowledge synthesis agent

The knowledge synthesis agent is responsible for knowledge integration, semantic interaction, and result presentation. It serves as the primary interface between the multi-agent system and the user. Leveraging the semantic understanding capabilities of LLMs, the agent supports domain-specific question answering. It can also automatically summarize intermediate and final design results and generate structured design reports. These functions improve the interpretability and consistency of the system.

During agent interactions, non-specialized design tasks and general semantic requests are typically routed to this agent. It handles task coordination and ensures unified information representation. In addition, it integrates outputs from different functional agents and produces coherent semantic summaries. The introduction of this agent enhances human-machine interaction and improves knowledge reuse and decision support in complex engineering design workflows.

## 4. Results and discussion

This section evaluates the comprehensive design capabilities of TurboAgent from four perspectives. First, the effectiveness of the task planning agent is assessed through queries of varying complexity. Second, the capabilities of individual functional agents are independently validated. Third, a complete end-to-end design task is conducted to demonstrate the overall performance of the proposed framework. Finally, a detailed cost analysis is performed, including token consumption and computational time.

### 4.1 Task planning agent validation

To systematically evaluate the effectiveness of the high-level task planning agent in natural language understanding, a set of validation scenarios with progressively increasing complexity is constructed. These scenarios are designed to assess the agent's capability to handle tasks ranging from basic instruction execution to complex engineering problem-solving. It should be noted that all experiments are conducted using the DeepSeek V3.2-Chat API.

The test cases cover typical engineering workflows, including both design generation and optimization analysis. In the basic linear planning scenario, the focus is on whether the agent can accurately extract explicit numerical requirements, such as mass flow rate and total pressure ratio, and construct a standard sequential workflow consisting of "generation-prediction" steps.

In the high-fidelity validation scenario, advanced numerical simulation is incorporated into the design process. This scenario evaluates whether the agent can correctly interpret the requirement of "high-fidelity validation" and automatically invoke external simulation tools to ensure physical consistency.

The task complexity is further extended to goal-driven iterative optimization, where the agent is required not only to perform performance prediction but also to make conditional decisions based on evaluation results. In this case, the agent autonomously constructs an iterative optimization workflow with feedback loops, demonstrating its capabilities for result-driven decision-making and adaptive planning.

Figure 5 illustrates the interaction processes and corresponding task workflows under different scenarios. The results show that the high-level task planning agent exhibits robust semantic understanding and logical reasoning capabilities. It effectively maps unstructured natural language inputs into structured graph-based workflows and adaptively adjusts planning strategies according to task requirements.

Specifically, for simple tasks, the agent prioritizes the performance prediction agent for rapid evaluation. For more complex scenarios involving high-fidelity validation or optimization, it automatically constructs workflows that incorporate multi-tool integration, conditional branching, and batch processing. These results demonstrate that the proposed framework achieves strong autonomy, robustness, and scalability in handling nonlinear and highly coupled turbomachinery design problems.

**Q1:** Design a compressor with a mass flow rate of 15 kg/s, a pressure ratio of 1.60, and an efficiency of 0.88. Provide three design schemes and perform fast performance validation.

📋 **Task Plan** — Multi-step workflow detected **Execution pipeline:** ① **Design Expert** (Blade Design) → ② **Evaluation Expert** (Performance Evaluation) > 💡 The user wants to design a compressor with specific performance targets (flow rate 15 kg/s, pressure ratio 1.60, efficiency 0.88). They request three design schemes and a fast performance validation. This is a multi-step request: first, generating multiple design schemes, then evaluating their performance …

**Q2:** Design a compressor with a mass flow rate of 15 kg/s, a pressure ratio of 1.60, and an efficiency of 0.88. Perform a fast performance validation first, and if the initial validation is satisfactory, proceed with physical aerodynamic and structural strength verification.

📋 **Task Plan** — Closed-loop workflow with conditional branching **Execution pipeline:** ① **Design Expert** (Blade Design) → ② **Evaluation Expert** (Performance Evaluation) → ③ **Simulation Expert** (CFD Simulation) → ④ **Simulation Expert** (FEA Simulation) **Conditional logic:** The user wants to design a compressor with specific performance targets (mass flow=15 kg/s, pressure ratio=1.60, efficiency=0.88). The request involves a multi-step verification process: 1) A 'fast performance validation' first. 2) If the initial validation is satisfactory, then proceed with physical aerodynamic (CFD) and structural strength verification …

**Q3:** Design a compressor with a mass flow rate of 15 kg/s, a pressure ratio of 1.60, and an efficiency of 0.88. Perform a fast performance validation, and if the results do not meet the targets, optimization is required.

📋 **Task Plan** — Closed-loop workflow with conditional branching **Execution pipeline:** ① **Design Expert** (Blade Design) → ② **Evaluation Expert** (Performance Evaluation) → ③ **Optimization Expert** (Design Optimization) **Conditional logic:** The user input is a complete workflow request in English: 1) Design a compressor with specific performance targets (mass flow=15 kg/s, pressure ratio=1.60, efficiency=0.88). 2) Perform a fast performance validation (this means using the ViT proxy model for evaluation, not CFD simulation). 3) If the results do not meet the targets, optimization is required …

Figure 5. Task planning results under different validation scenarios.

## 4.2 Functional agent validation

The effectiveness of individual components in the proposed multi-agent system is evaluated through single-task validation. Each agent is independently assessed to examine its capability in achieving the corresponding design objectives.

### 4.2.1 Generative design task validation

To evaluate the responsiveness of the generative design agent to user-defined requirements and its capability for design diversity, a set of experiments was conducted using 100 randomly generated performance targets that are not included in the training dataset. Each target specifies $\dot{m}$, $\pi$, and $\eta$. Among these, three representative objectives are selected for detailed analysis, including maximum $\dot{m}$, maximum $\pi$, and maximum $\eta$. To rigorously assess design diversity, the agent generated 10 candidate geometries for each of the three selected objectives, yielding a total of 130 design samples across the entire experiment.

Taking a representative case as an example, Fig. 6 illustrates the interaction process between the user and the agent. In the initial query, only mass flow rate and total pressure ratio are specified, while the isentropic efficiency target is omitted. In response, the agent proactively requests the missing efficiency requirement to ensure a complete specification of the design objective. This behavior demonstrates its capability in intelligent requirement extraction and semantic completion.

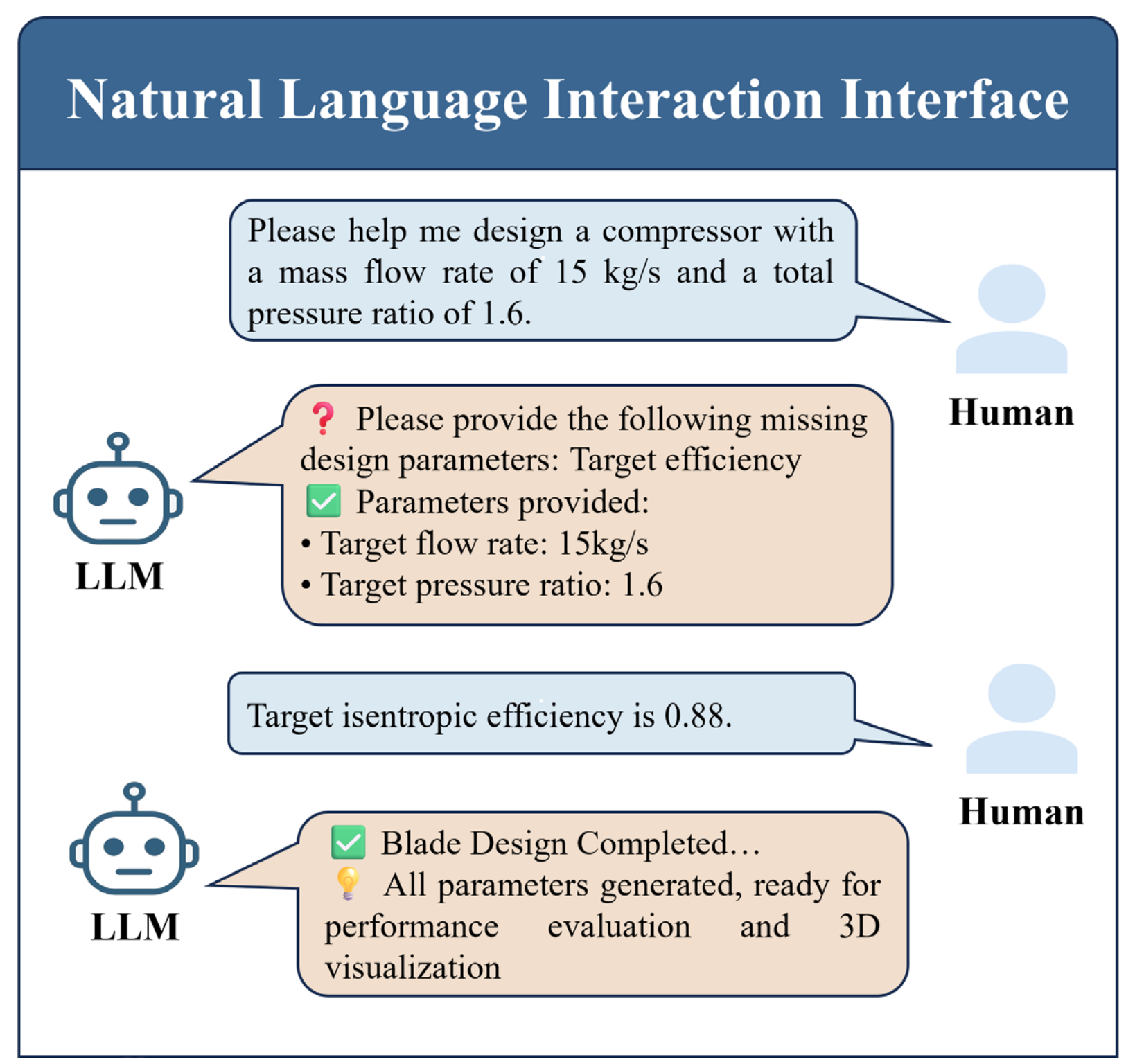

Figure 6. Interaction process between the user and the generative design agent.

Figure 7 presents the generated results for the three representative objectives. For each target, the agent completes generative inverse design, parameterized modeling, and front-end visualization within one minute, while providing multiple candidate solutions. The results indicate that the generated blade geometries exhibit smooth profiles and satisfy geometric constraints. For a given design objective, the agent produces multiple distinct geometric configurations, demonstrating strong design diversity. This diversity provides a rich set of candidate solutions for subsequent evaluation and selection. Such diversity is particularly valuable in engineering applications, as it enables designers to select optimal solutions from multiple feasible candidates. In multidisciplinary design scenarios, where additional constraints may render certain designs infeasible, design diversity enhances the robustness and adaptability of the overall design process.

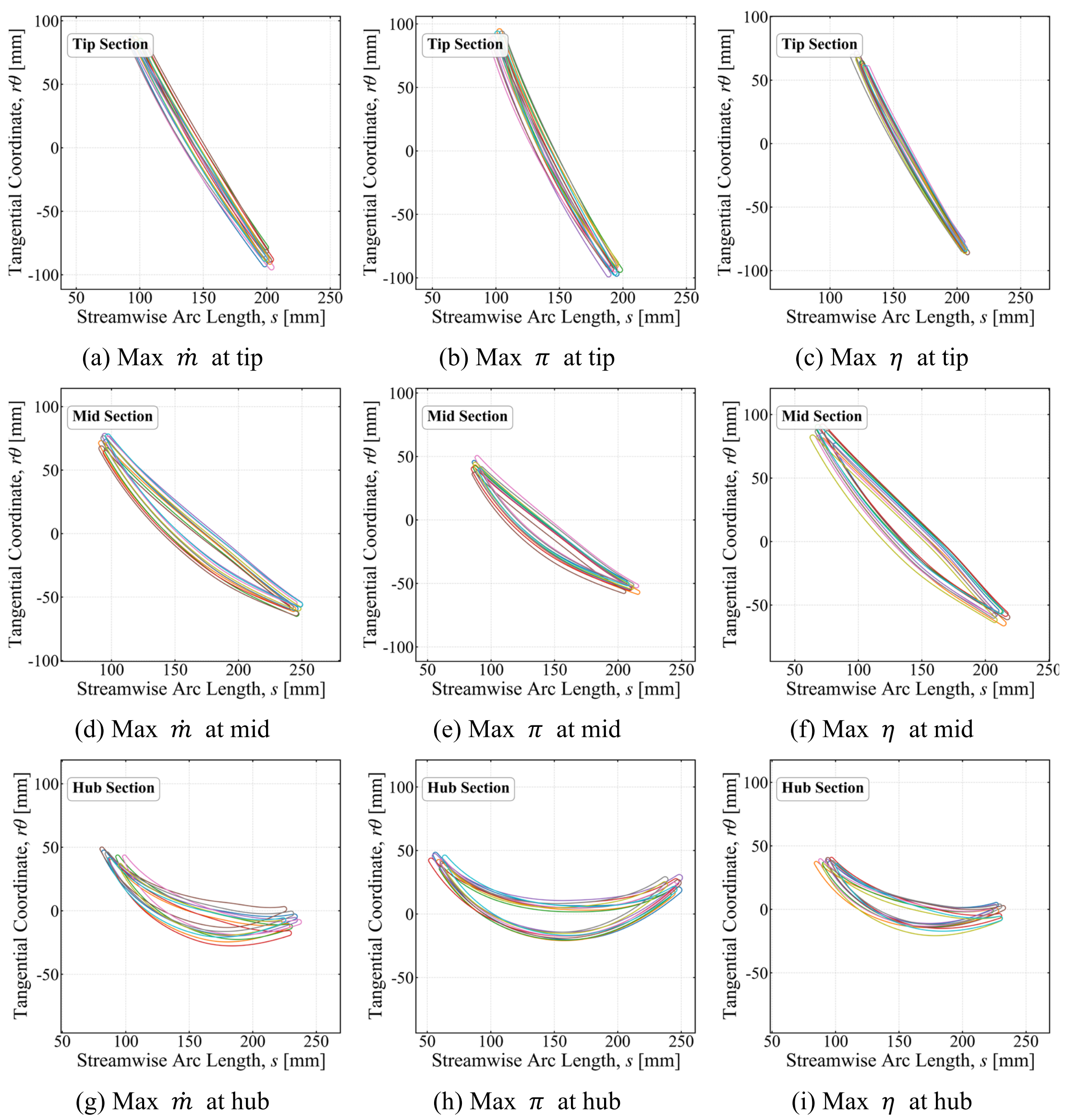

(a) Max $\dot{m}$ at tip (b) Max $\pi$ at tip (c) Max $\eta$ at tip

(d) Max $\dot{m}$ at mid (e) Max $\pi$ at mid (f) Max $\eta$ at mid

(g) Max $\dot{m}$ at hub (h) Max $\pi$ at hub (i) Max $\eta$ at hub

Figure 7. Generated blade geometries under different design objectives.

### 4.2.2 Performance prediction task validation

The objective of the performance prediction agent is to replace computationally expensive CFD solvers with millisecond-level design evaluation. In this section, the agent is applied to rapidly assess the 130 candidate designs generated in Section 4.2.1.

**Figure 8** presents the prediction results, where the horizontal axis represents the target design values and the vertical axis corresponds to the predicted values. The coefficient of determination ($R^2$) is used to quantify the proportion of variance explained by the model, where values closer to 1 indicate a stronger predictive capability. In addition, the normalized root mean square error (nRMSE) and mean absolute error (MAE) are employed to evaluate prediction accuracy. The results indicate that, as evaluated by the performance prediction agent, the geometries generated by the design agent exhibit strong consistency with the design targets. Specifically, $R^2$ values exceed 0.99 for all performance metrics, and nRMSE values remain below 2%. These results, to some extent, reflect the accuracy of the generative design agent. Nevertheless, the final design accuracy still requires further validation through high-fidelity simulations.

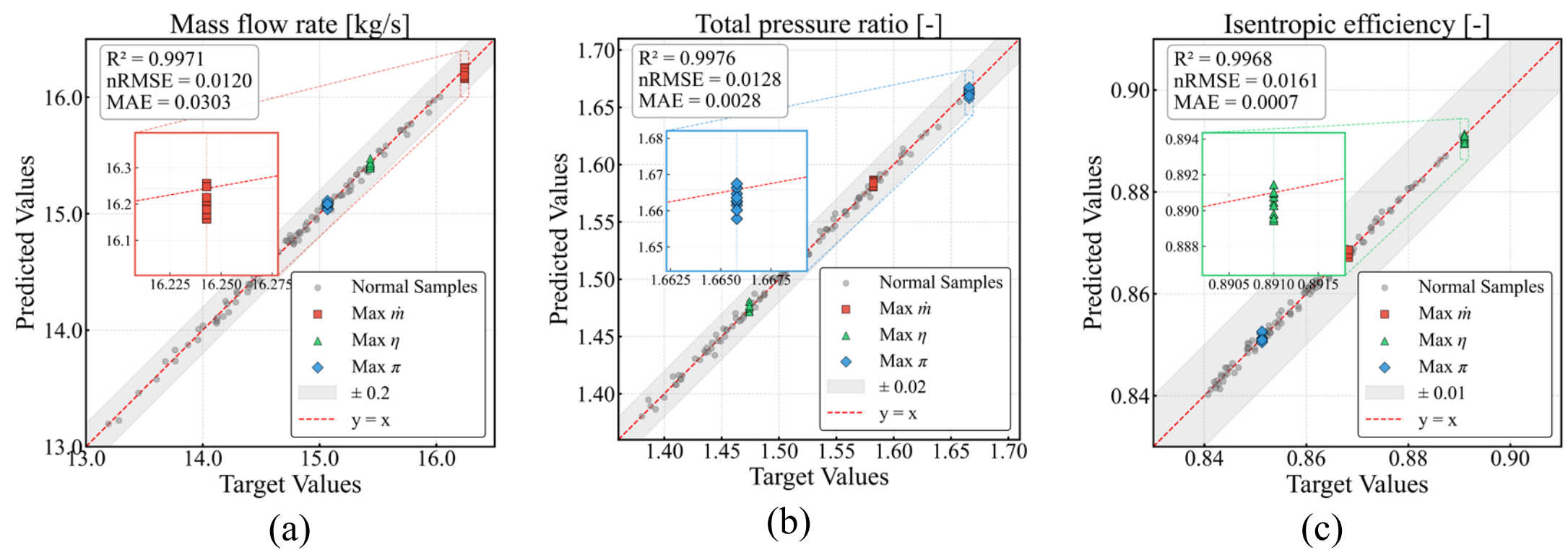

Figure 8. Predicted versus target performance of inverse designs.

In addition to batch evaluation, the agent supports interactive design parameter modification through natural language input. Designers can specify analysis tasks using simple instructions, such as “investigate the influence of the blade tip chord length on performance.” The agent then automatically updates the corresponding geometric parameters and outputs new performance predictions. Fig. 9 illustrates this interactive process. The near-instantaneous feedback enables efficient exploration of local design variations and facilitates rapid identification of promising optimization directions, significantly improving the efficiency of the design iteration process.

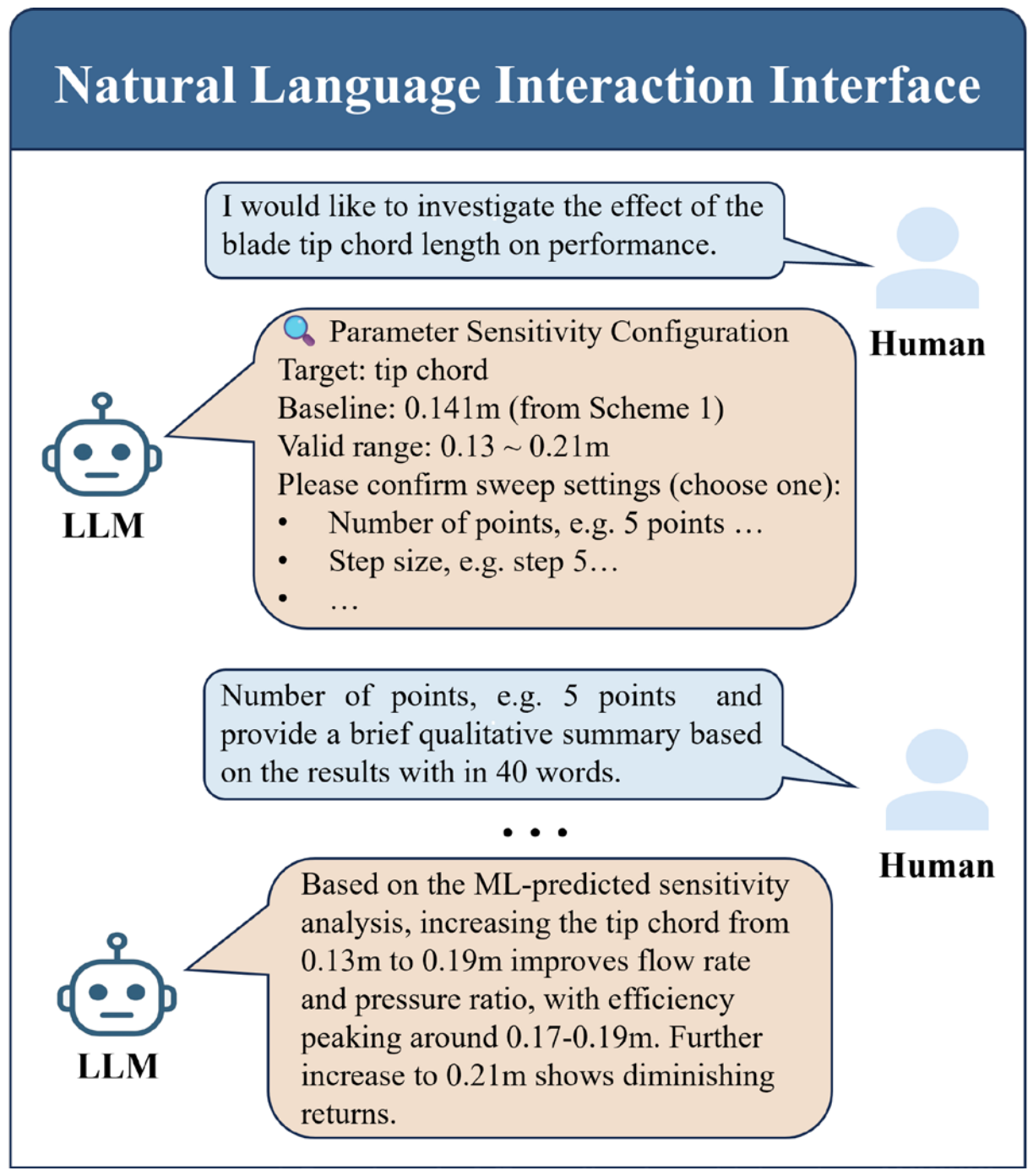

Figure 9. Interactive design modification and performance prediction.

### 4.2.3 Physics simulation task validation

To verify the physical fidelity of the generated designs and evaluate the accuracy of the performance prediction agent, the physics validation agent performs automated CFD simulations on the generated candidates. It should be noted that the agent supports parallel computation based on available CPU cores, enabling the efficient evaluation of multiple designs.

Figure 10 illustrates an example of user interaction for performing physical simulation through natural language input. Once the agent identifies the user's request, it automatically constructs a workflow that includes mesh generation, preprocessing, and solver execution. These processes are implemented through interactions between the agent and CFD/FEA software APIs. During the simulation, convergence histories of key performance indicators are visualized in the front-end interface, allowing real-time monitoring of solver progress. After completion, the agent prompts the user for post-processing. Upon confirmation, it automatically generates a script compatible with ANSYS CFX-Post and extracts key physical quantities from the simulation results. The processed results are directly visualized in the front-end interface, enabling rapid interpretation of simulation outcomes.

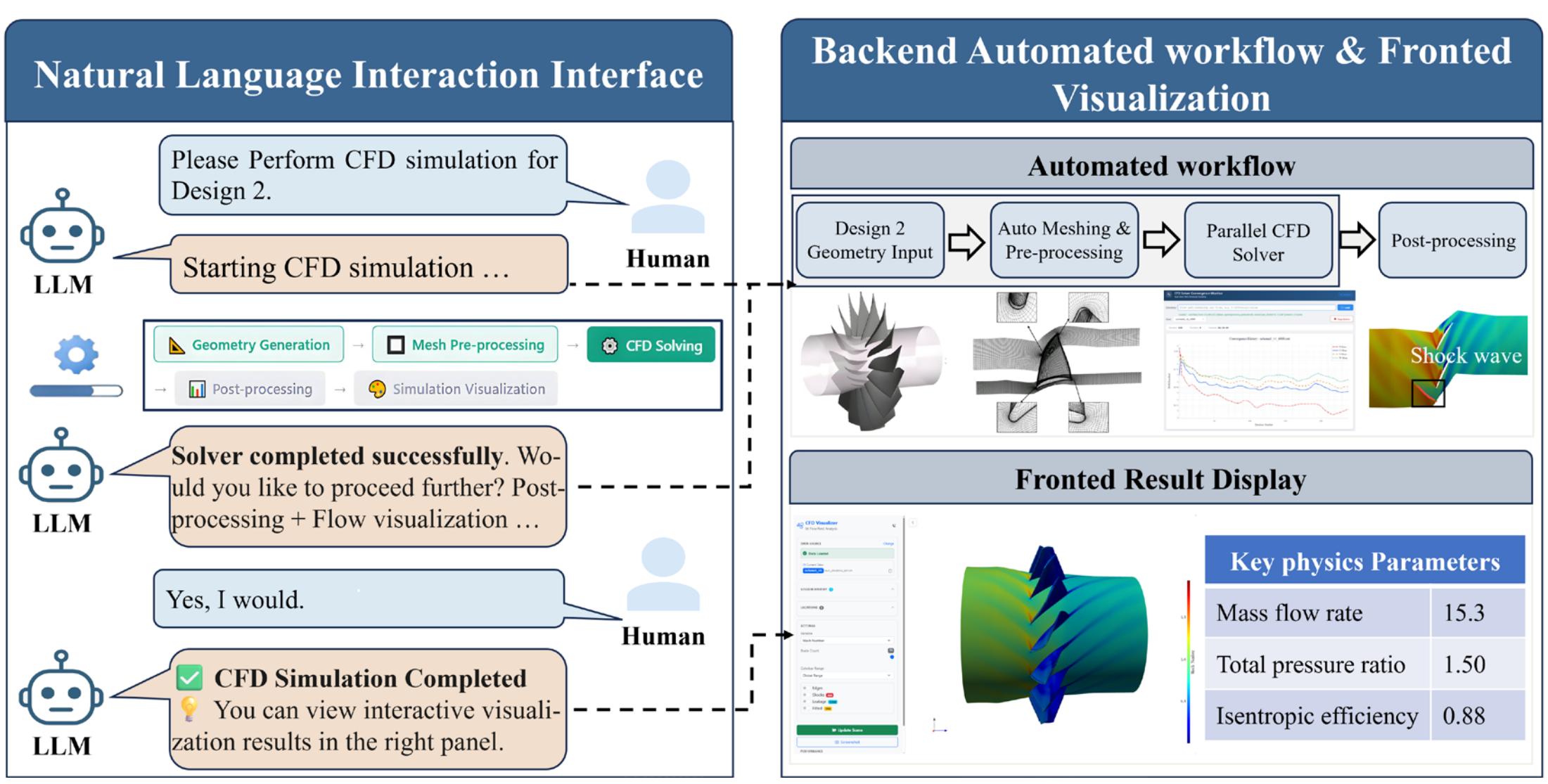

Figure 10. Interactive physical simulation workflow driven by natural language input.

Figure 11 presents the CFD results of 130 generated geometries along with a comparative analysis. The figure is divided into two columns. The left column compares the design targets with CFD simulation results, where the horizontal axis represents the target values and the vertical axis represents the CFD results. The right column compares the CFD results with the predictions from the performance prediction agent, where the horizontal axis denotes the CFD values and the vertical axis denotes the predicted values. The color of each data point indicates the absolute relative error (ARE) with respect to the design targets. It should be noted that some simulations failed to converge. As a result, 124 valid cases were obtained for analysis, corresponding to an overall design success rate of approximately 95%.

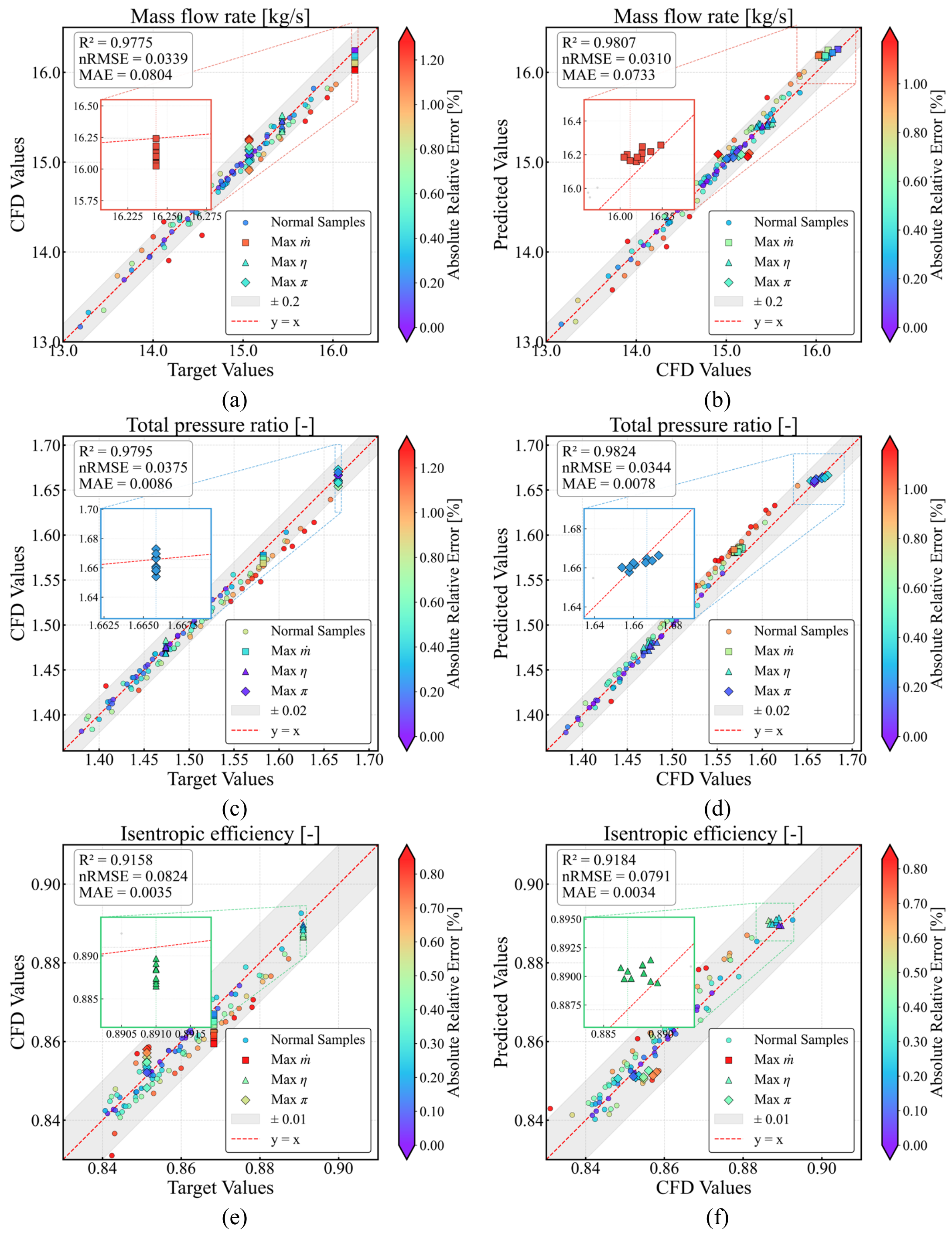

Figure 11. Comparison of design targets, CFD results, and predicted performance.

From the left column, a strong agreement between CFD results and design targets can be observed. The $R^2$ values reach 0.9775 for $\dot{m}$, 0.9795 for $\pi$, and 0.9158 for $\eta$. The nRMSE values for all performance metrics remain below 9%. From the right column, the CFD results also show high consistency with the predictions of the performance prediction agent. The $R^2$ values for $\dot{m}$, $\pi$, and $\eta$ are 0.9807, 0.9824, and 0.9184, respectively, with nRMSE values below 8% for all metrics.

Overall, these results demonstrate that the compressor blade geometries generated by the cDDPM can effectively approximate the target aerodynamic performance. Meanwhile, the performance prediction model achieves high prediction accuracy. Together, these findings validate the effectiveness of the proposed generative model-based multi-agent framework.

In addition, the agent integrates structural validation capabilities. Designers can modify material properties or operating conditions through natural language instructions (e.g., “change the material from stainless steel to Ti-6Al-4V and evaluate stress”). The agent then invokes a FEM solver to compute the equivalent stress distribution of the blade at maximum rotational speed (as shown in Fig. 12). This functionality ensures that the generated designs not only meet aerodynamic requirements but also satisfy structural strength constraints, which is critical for practical engineering applications, especially in the early design stage.

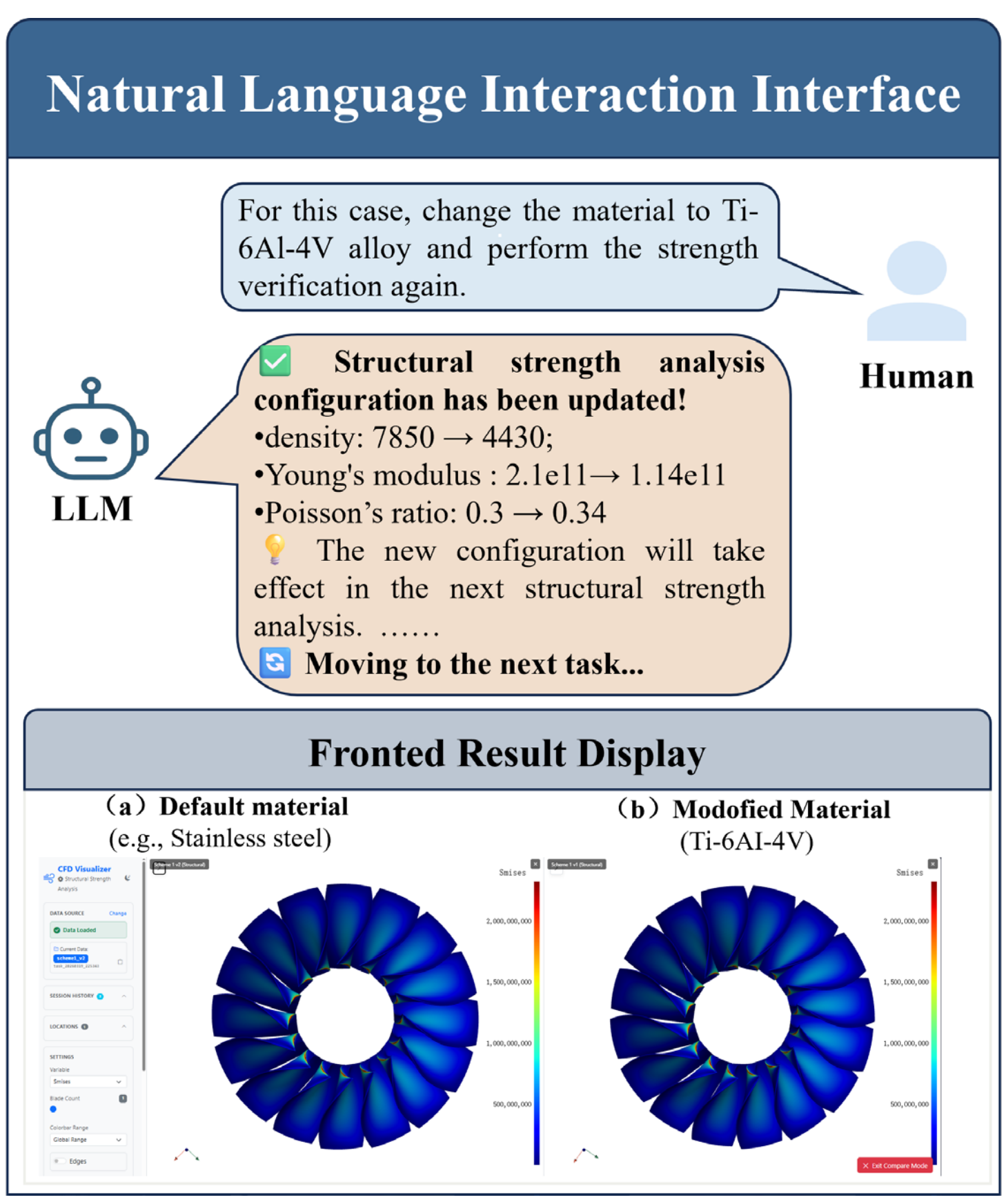

Figure 12. Interactive structural validation and FEM-predicted stress distribution of the blade.

### 4.2.4 Optimization task validation

In the performance optimization task, an initial design is specified with a design point of $\dot{m}$ 15 kg/s, $\pi$ 1.6, and $\eta$ 0.87. After performance prediction, the user can further refine the optimization objectives through natural language input. For example, the user may specify: “optimize the total pressure ratio and isentropic efficiency while maintaining the design-point mass flow rate.” The task planning agent accurately interprets the optimization intent and automatically triggers the subsequent optimization workflow.

During the optimization stage, the optimization agent integrates multiple algorithms, including an LLM-driven optimization algorithm, the GA, and PSO. By interacting with the surrogate model, the agent iteratively generates improved design candidates. Fig. 13 presents the interaction process and the corresponding optimization results. The results show that the task planning agent correctly parses the user input, transfers the optimization objectives to the optimization agent, and automatically constructs the corresponding reward function.

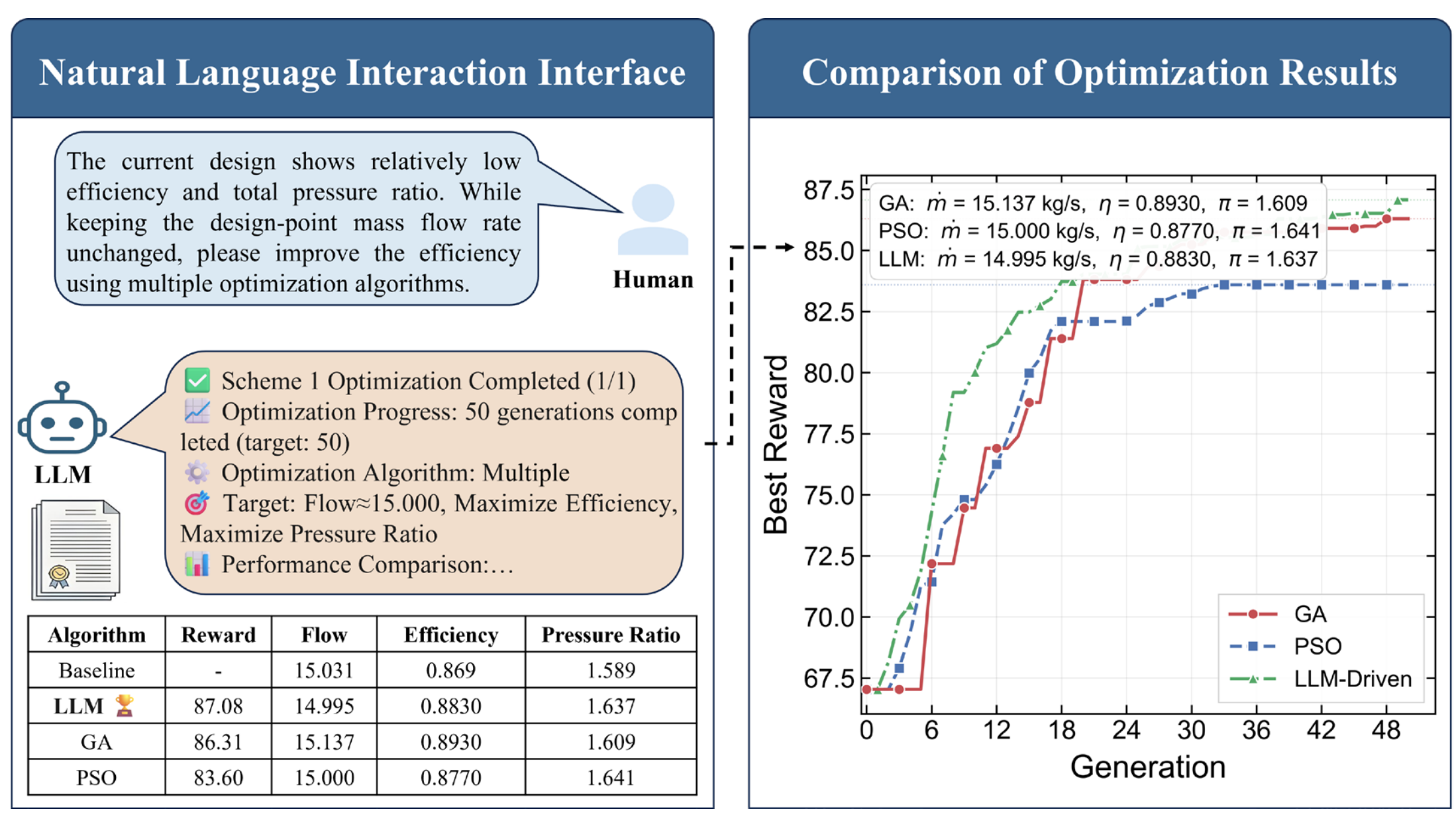

| Algorithm | Reward | Flow | Efficiency | Pressure Ratio |
|---|---|---|---|---|
| Baseline | - | 15.031 | 0.869 | 1.589 |
| **LLM** 🏆 | 87.08 | 14.995 | 0.8830 | 1.637 |
| GA | 86.31 | 15.137 | 0.8930 | 1.609 |
| PSO | 83.60 | 15.000 | 0.8770 | 1.641 |

Figure 13. Optimization process and convergence results using different optimization methods.

A comparison of convergence behaviors among different optimization methods shows that the LLM-driven approach exhibits clear advantages in both flexibility and optimization efficiency. It achieves faster convergence in the early stages compared with GA and PSO, and attains the highest final reward value. The optimized design shows notable improvements over the initial configuration, with the isentropic efficiency increased by 1.61% and total pressure ratio improved by 3.02%.

It should be noted that the performance of the LLM-based optimization method depends on the capabilities of the underlying model. Variations in model scale and architecture may lead to differences in optimization performance. However, the proposed optimization agent supports the flexible integration of multiple algorithms, ensuring robustness and adaptability across different optimization tasks.

After the optimization process, the agent automatically generates a detailed report. This report summarizes the changes in design parameters, geometric modifications, and performance improvements before and after optimization. It also provides a comparative analysis of different optimization methods, highlighting their respective advantages and limitations. This functionality helps users better understand the applicability of different optimization strategies in practical engineering problems and supports informed decision-making in various design scenarios.

### 4.3 Multi-task workflow validation

To validate the end-to-end execution capability of the proposed multi-agent framework in complex engineering design tasks, a complete compressor design case was constructed. The design objectives were a mass flow rate of 15.2 kg/s, an isentropic efficiency of 0.88, and a total pressure ratio of 1.62.

Under this task, the system autonomously completes the entire design process driven by natural language, from initial concept generation to final design selection. First, the generative design agent produces 10 candidate geometries based on the given performance targets. These candidates are rapidly evaluated and parameterized using the performance prediction agent, and the system automatically determines whether further optimization is required. Subsequently, the physics validation agent performs high-fidelity aerodynamic and structural analyses at the design point for selected promising candidates, ensuring physical consistency.

Based on these results, the high-level task planning agent makes comprehensive decisions by integrating simulation outcomes with design objectives. Four representative optimal candidates are selected. Further off-design analyses are then triggered, including performance map generation and stability evaluation. Finally, the optimal design is determined through multi-objective trade-off analysis. The overall workflow and results are illustrated in Fig. 14.

Figure 15 further presents representative human-machine interaction examples during the design process, including user natural language inputs and key reasoning and decision outputs from TurboAgent. The results demonstrate that the agents can systematically evaluate candidate designs by jointly considering aerodynamic performance, structural integrity, and stability margin. Among all candidates, Scheme 9 is identified as the final optimal design. It is worth emphasizing that the entire workflow operates without explicit human intervention, achieving seamless task transitions and dynamic adjustments across multiple stages. Meanwhile, the system provides full-process visualization through a front-end interface and supports human-machine collaboration via natural language interaction. In addition, the LLM can generate structured summaries of the design process and automatically produce design reports, highlighting its capability in reasoning and knowledge integration for complex engineering problems.

Overall, the multi-task workflow validation results demonstrate that the proposed framework can achieve a fully automated closed-loop compressor design process. Compared with conventional methods, the framework requires only natural language input to complete complex design tasks. It shows significant advantages in design efficiency, solution exploration capability, and intelligent decision-making, highlighting its strong potential for applications in complex turbomachinery aerodynamic design.

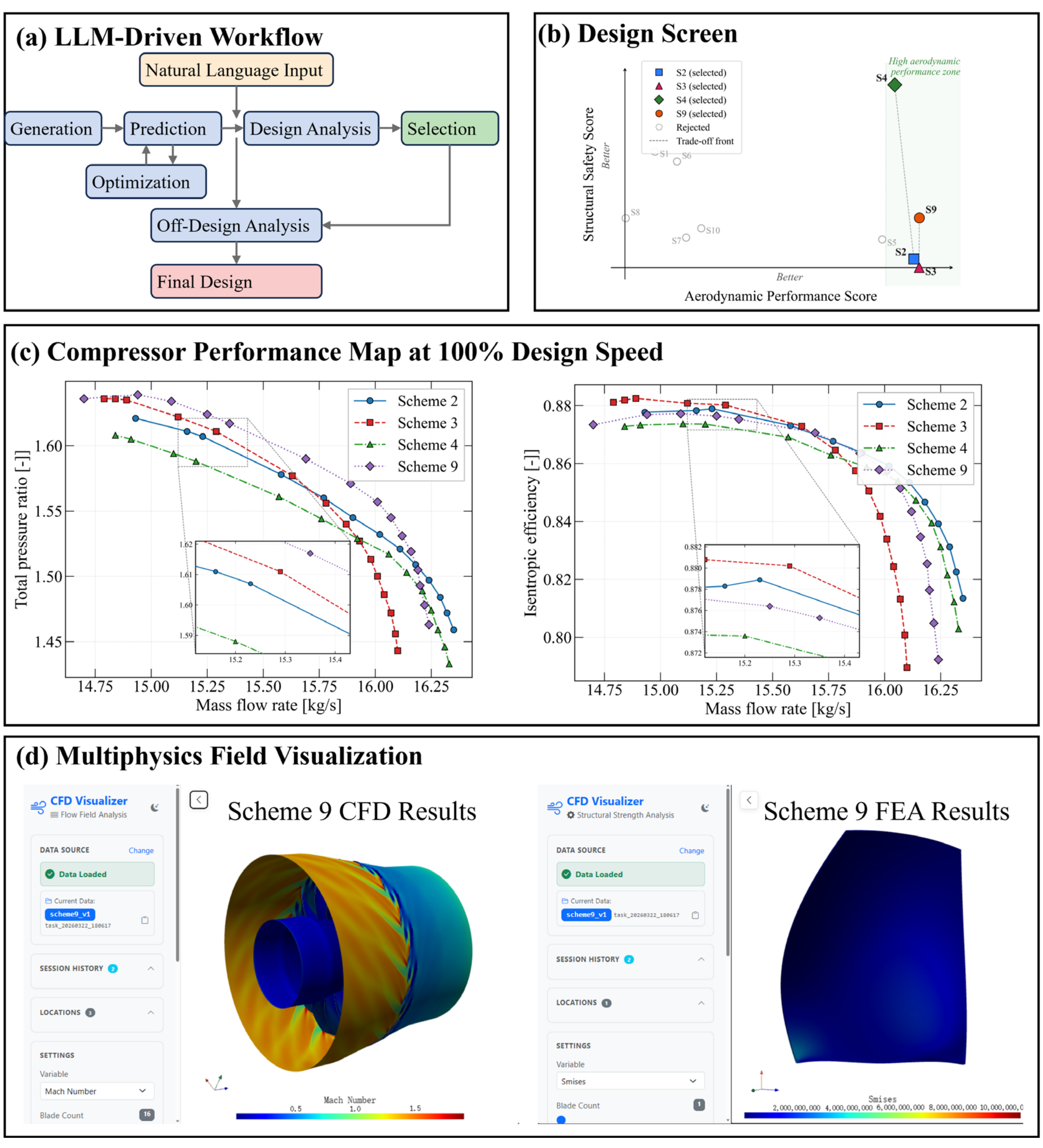

Figure 14. Overall workflow and results of the design process.

**Q1:** Please help me complete a compressor design task. The required design-point performance is a mass flow rate of 15.2 kg/s, a total pressure ratio of 1.62, and an efficiency of 0.88. Please generate 10 design schemes and perform performance evaluation and blade geometry visualization. If necessary, carry out further optimization; then carry out CFD aerodynamic simulations and structural strength simulations for all 10 schemes for validation.

✅ **Multi-Task Sequence Completed**

Completed tasks: Design → Evaluation →Optimization →Visualization → Simulation (5 total)

💡 Detailed results for each task are shown above. You can continue asking questions or start new tasks.

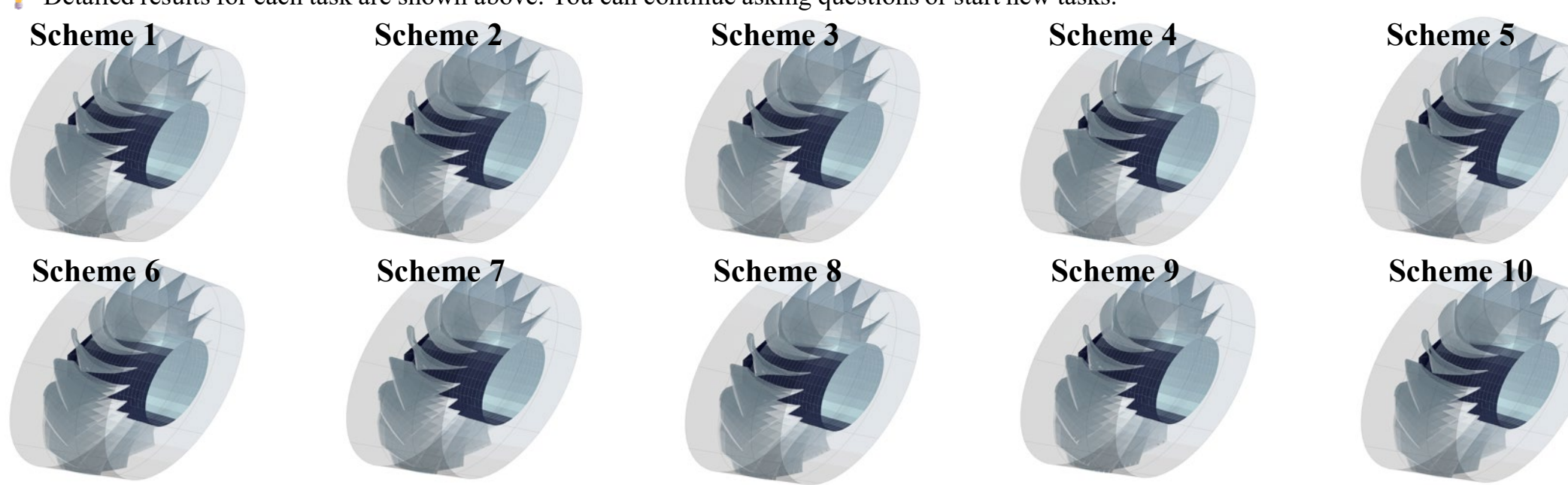

**Q2:** Based on the CFD and structural strength simulation results of the above designs, select the four best design schemes from them. Keep the answer short — one summary table and a 2-3 sentence conclusion.

✅ **Compressor Blade Design Scheme Selection Analysis:**

Based on the actual CFD and structural strength simulation results (not ML predictions), I have analyzed all 10 design schemes against your performance targets and structural requirements. Here are the four best-performing schemes…

| Scheme | Mass Flow (kg/s) | Efficiency | Pressure Ratio | Max Stress (GPa) | Key Assessment |
|---|---|---|---|---|---|
| Scheme 3 | 15.29 | 0.8802 | 1.611 | 15.30 | Best performer. Closest to all targets, highest efficiency. |
| Scheme 2 | 15.23 | 0.8789 | 1.607 | 14.36 | Excellent performance, very close to Scheme 3. |
| Scheme 9 | 15.35 | 0.8753 | 1.617 | 10.58 | Highest flow & pressure ratio, good efficiency. |
| Scheme 4 | 15.20 | 0.8736 | 1.588 | 3.95 | Best structural integrity (lowest stress), meets flow target. |
| Target | 15.20 | 0.8800 | 1.620 | | - |

**Conclusion:** Schemes 3 and 2 are the top aerodynamic performers, nearly meeting all design targets. Scheme 9 offers the highest total pressure ratio, while Scheme 4 is the most structurally robust. All four exceed the target mass flow, but efficiencies are slightly below the 0.88 goal, indicating a potential area for further optimization

**Q3:** I need to calculate the performance maps of these four recommended schemes at 100% rotational speed. Then, based on the design targets, evaluate Schemes 2, 3, 4, and 9 using: (1) design-point CFD performance deviation from targets, (2) off-design operating range and stability margin from performance maps, (3) structural integrity (max Von Mises stress). Present the comparison in a single table covering all three dimensions, then recommend the final scheme with a quantitative justification (3-5 sentences). Do not repeat raw operating-point data..

✅ **Comprehensive Evaluation of Candidate Schemes**

| Scheme | Design-Point CFD Performance (vs. Target) | Off-Design Stability & Range | Structural Integrity |
|---|---|---|---|
| Target | Flow: 15.20 kg/s, η: 0.880, PR: 1.620 | Wide operating range | Below material yield strength |
| Scheme 2 | Flow: +0.2% (15.23 kg/s)<br>Efficiency: -0.13% (0.8789)<br>PR: -0.80% (1.607) | Surge Margin: 2.90%<br>Operating range: 14.93-16.35 kg/s<br>Limited stability margin | $1.44\times10^{10}$ Pa<br>Very high stress concentration |
| Scheme 3 | Flow: +0.6% (15.29 kg/s)<br>Efficiency: +0.02% (0.8802)<br>PR: -0.56% (1.611) | Surge Margin: 4.98%<br>Operating range: 14.79-16.10 kg/s<br>Best stability among candidates | $1.53\times10^{10}$ Pa<br>Highest stress, requires careful material selection |
| Scheme 4 | Flow: 0.0% (15.20 kg/s)<br>Efficiency: -0.73% (0.8736)<br>PR: -1.98% (1.588) | Surge Margin: 3.72%<br>Operating range: 14.84-16.33 kg/s<br>Moderate stability | $3.95\times10^{9}$ Pa<br>Best structural integrity by far |
| **Scheme 9** | **Flow: +1.0% (15.35 kg/s)**<br>**Efficiency: -0.53% (0.8753)**<br>**PR: -0.19% (1.617)** | **Surge Margin: 5.65%**<br>**Operating range: 14.70-16.24 kg/s**<br>**Widest stability margin** | **$1.06\times10^{10}$ Pa**<br>**High but manageable stress** |

🏆 **Final Recommendation: Scheme 9**

Quantitative Justification:

Scheme 9 provides the best balance of aerodynamic performance and operational safety. It achieves the closest pressure ratio to target (-0.19% deviation, best among all candidates) while delivering the highest mass flow rate (+1.0%). Most critically, it offers the widest surge margin at 5.65%, which is 90% higher than Scheme 2 and provides significantly better protection against stall during off-design operation. Although its efficiency is 0.53% below target, this is a reasonable trade-off for the substantial gain in stability. Structurally, while its stress ($1.06\times10^{10}$ Pa) is high, it is 31% lower than Scheme 3 and remains within design limits for advanced compressor alloys with appropriate safety factors.

Figure 15. Human-machine interaction in the LLM-Driven design process.

### 4.4 Cost analysis

To further demonstrate the advantages of the proposed framework in improving efficiency and reducing cost, this section analyzes the computational cost of the framework, as illustrated in Fig. 16.

The overall computational overhead mainly comes from two sources: token consumption during LLM inference and interaction, and computational resources (e.g., CPU cores) required for high-fidelity simulations. This study focuses on the LLM-related cost. Considering the variations in performance and cost across different LLM architectures, the DeepSeek V3.2-Chat model is selected as a representative example.

For a basic single design task, Fig. 16 shows the token consumption of various subtasks during execution. In multi-task scenarios, the total cost can be approximated as the linear accumulation of single-task consumption. Report generation tasks are not included in this analysis, as their token usage varies significantly depending on the level of detail. The results indicate that, except for LLM-based optimization tasks, the token consumption of other tasks remains relatively low. In contrast, the LLM optimization process involves iterative interactions, leading to significantly higher token usage that increases approximately linearly with the number of iterations. Overall, the total token consumption for completing the full workflow, from design generation to simulation-based validation, is approximately 80,000 to 100,000 tokens.

From a time-cost perspective, the main computational burden lies in the high-fidelity simulation stage. Although the framework automates design generation, pre-processing, and post-processing, solving the Navier-Stokes equations still requires considerable computational time. However, this cost can be effectively reduced using high-performance computing resources. Under parallel execution with 30 CPU cores, the simulation time for a single operating condition is approximately 30 minutes. Consequently, the entire workflow can be completed within about 30 minutes, enabling an efficient closed-loop automated design process. Compared with conventional design workflows that typically require several weeks, the proposed framework achieves an order-of-magnitude reduction in design time, highlighting its significant potential for efficient engineering design.

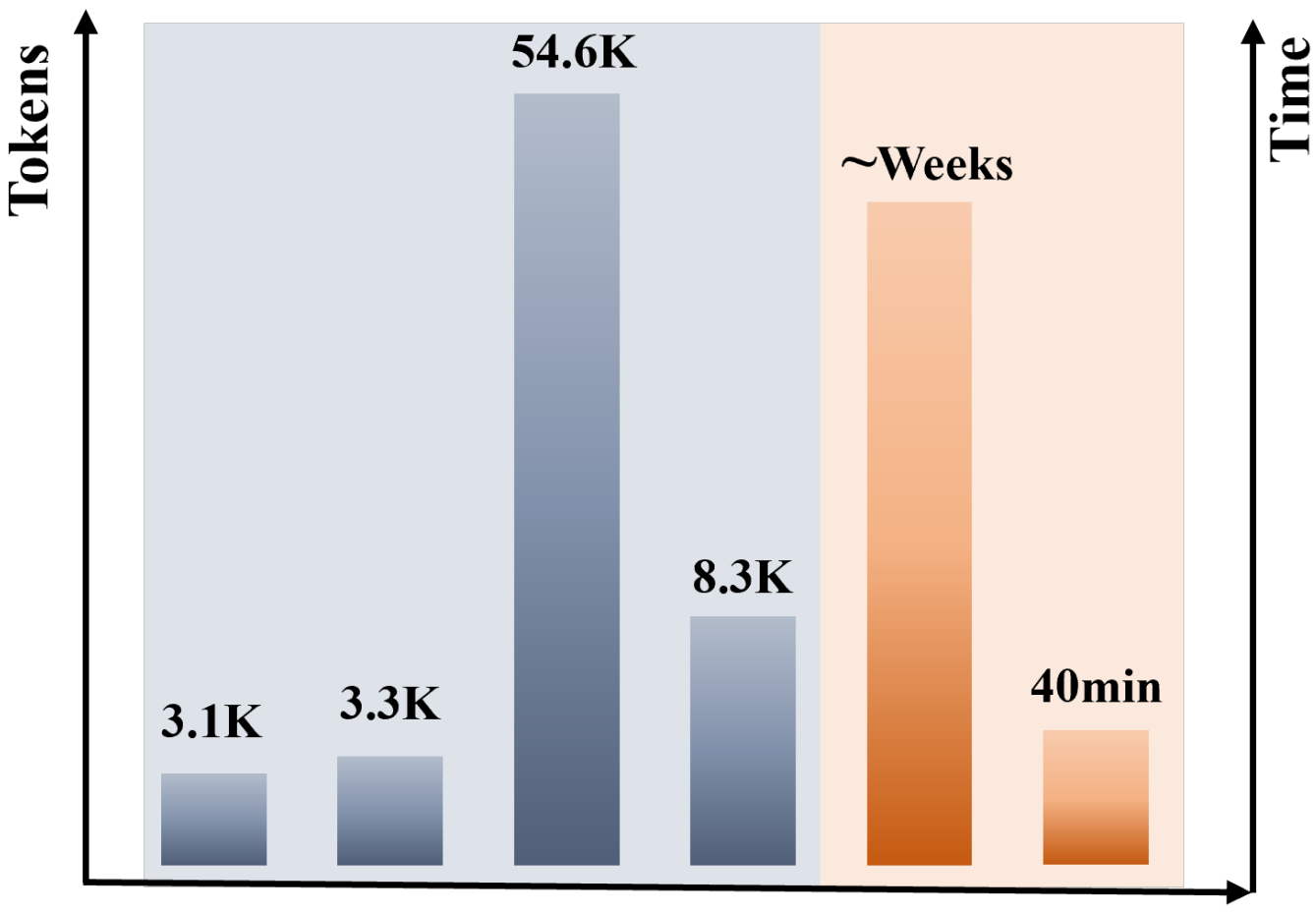

Figure 16. Cost analysis of TurboAgent.

## 5. Conclusion

This study proposes TurboAgent, an LLM-driven autonomous multi-agent framework for turbomachinery aerodynamic design and optimization. The proposed framework was systematically validated using a transonic single-rotor compressor. The results show that the framework enables an end-to-end automated workflow, from requirement parsing to physical validation, and provides an efficient and scalable solution for complex aerodynamic design tasks. The main conclusions and contributions are summarized as follows:

(1) A data-driven iterative design loop is established to improve design efficiency. Based on cDDPM, the generative design agent can rapidly produce high-quality and highly diverse 3D compressor blade configurations tailored to multidimensional performance targets. Additionally, the Transformer-based performance prediction agent achieves millisecond-level evaluation. Its prediction accuracies ($R^2$) for $m$, $\pi$, and $\eta$ reach 0.9807, 0.9824, and 0.9184, respectively, while all nRMSE values remain below 8%. These results demonstrate that the main design loop can efficiently generate and screen preliminary candidate designs, thereby providing effective inputs for subsequent optimization and physical validation.

(2) The meta-prompt-driven LLM optimizer can further enhance design quality. The optimization agent incorporates an LLM-based aerodynamic optimization method alongside traditional algorithms. Compared with the GA and PSO, the LLM-based optimization method shows clear advantages in convergence speed and optimization efficiency. It improves the $\eta$ by 1.61% and the $\pi$ by 3.02%, respectively, relative to the initial design. This indicates that the LLM can deeply understand the semantics of design targets and constraints, adaptively adjusting search strategies without the need to explicitly construct evolutionary operators.

(3) Automated multi-physics high-fidelity validation effectively supports final engineering decision-making. The physics validation agent enables automated integration with commercial CFD solvers and FEA modules. Driven by natural language commands, it can automatically execute mesh generation, boundary condition configuration, parallel solving, and result post-processing, while also supporting material replacement and stress verification. The high-fidelity validation results show good agreement between the design targets and final simulation results, with $R^2 > 0.91$ and nRMSE $< 8\%$. This confirms that the framework can support final design decisions while maintaining physical fidelity.

Overall, TurboAgent not only integrates multiple specialized agent modules but also establishes a complete autonomous closed-loop design workflow encompassing requirement understanding, scheme generation, performance prediction, intelligent optimization, and physical validation. It provides an efficient and scalable new paradigm for turbomachinery aerodynamic design. More importantly, it promotes a shift from conventional experience-based and simulation-driven iterative design toward an autonomous workflow characterized by unified LLM planning and collaborative multi-agent execution. Future research should focus on mitigating training data dependency, improving generalization capabilities across different configurations and operating conditions, and enhancing autonomous decision-making and interpretability in more complex scenarios.

## ACKNOWLEDGEMENTS

The authors are grateful for the support from the National Natural Science Foundation of China (Grant Nos. 52506059, 62595792, 52425603, 117GJHZ2023043GC).

## DATA AVAILABILITY STATEMENTS

The data that support the findings of this study are available from the corresponding author upon reasonable request.

## Appendix A: Core Prompts and Project Management of the Multi-Agent System

### A.1 Prompts

This appendix presents the core prompts used to drive the multi-agent system (MAS). These prompts are injected into a general large language model (DeepSeek-V3.2) at runtime. They enable dynamic planning and decision-making for engineering design tasks. The system does not rely on model fine-tuning. Instead, all domain knowledge is encoded and constrained through prompt engineering. This approach allows controllable modeling and execution of complex design workflows. The complete set of prompts, covering all six types of agents, is publicly available at: https://github.com/wuyueteng/TurboAgent-prompts

### A.2 Project Structure

Figure 17 illustrates the file organization in the design workflow. It includes both commercial software dependencies and files generated during the design process. To improve scalability and maintainability, a hierarchical storage strategy is adopted. Different types of data-such as geometry files, mesh files, simulation results, and logs—are organized separately. This structure supports efficient data sharing among agents. It also facilitates traceability and reproducibility of the design process. As a result, the overall engineering implementation becomes more standardized and reliable.

```
geometry_generate/ ← Root Directory
|
├── csv_files/ ← [Design Parameters]
| └── task_{YYYYMMDD_HHMMSS}/ Indexed by: (task_id)
| ├── scheme1_v1.csv 21D blade geometry parameters
| ├── scheme1_v2.csv (after user modification)
| └── scheme2_v1.csv
|
├── dat_files/ ← [2D Blade Profiles]
| └── task_{timestamp}/ Indexed by: (task_id, scheme, version)
| └── scheme{i}_v{v}/
| └── scheme{i}_v{v}BladeIn.dat
/
├── geomturbo_files/ ← [3D Geometry]
/ └── task{timestamp}/ Indexed by: (task_id, scheme, version)
| └── scheme{i}_v{v}/
| ├── scheme{i}_v{v}.geomTurbo Original scale → CFD simulation
| └── scheme{i}_v{v}scaled.geomTurbo Scaled → Structural analysis
/
├── cfd_results/ ← [CFD Simulation Results]
/ └── task{timestamp}/ Indexed by: (task_id, scheme, version)
| └── scheme{i}v{v}/
/ ├── mesh/ CGNS mesh + DEF definition files
/ ├── solve/ .res + .out solver output
/ └── post/
/ └── flow_field/
/ ├── span{xx}flow.csv Spanwise flow field data
/ └── performance.txt Aerodynamic performance metrics
/
├── structural_results/ ← [Structural Analysis Results]
/ └── task{timestamp}/ Indexed by: (task_id, scheme, version)
| └── scheme{i}_v{v}/
| ├── *.plt Stress distribution (mesh data)
| └── performance.dat Stress summary: Sx~Sxz + Seq(Smises)
|
├── CFD_dependent_files/ ← [Solver Templates]
| ├── *.ccl, *.pre CFX-Pre configuration templates
| └── *.igg NUMECA mesh generation templates
|
└── structure_strength_depend_files/ ← [Structural Templates]
├── strBC.dat Boundary conditions (ω = 22080 rpm)
├── strmat.dat Material properties (ρ, E, ν)
└── data.dat, MeshPar.dat FEM mesh parameters
```

Figure 17. System directory architecture of TurboAgent.